\documentclass{article} 
\usepackage{iclr2021_conference,times}

\usepackage{amsmath,amsfonts,bm}

\def\eqref#1{equation~\ref{#1}}

\def\1{\bm{1}}

\DeclareMathAlphabet{\mathsfit}{\encodingdefault}{\sfdefault}{m}{sl}
\SetMathAlphabet{\mathsfit}{bold}{\encodingdefault}{\sfdefault}{bx}{n}

\usepackage{caption} 
\usepackage{hyperref}
\usepackage{url}
\usepackage{tabularx}

\usepackage[utf8]{inputenc} 
\usepackage[T1]{fontenc}    
\usepackage{booktabs}       
\usepackage{amsfonts}       
\usepackage{nicefrac}       
\usepackage{microtype}      
\usepackage{lipsum}
\usepackage{times}
\usepackage{helvet}
\usepackage{courier}
\usepackage{graphicx}
\usepackage{subcaption}
\usepackage{mathrsfs}  
\usepackage{amssymb}
\usepackage{amsmath}
\usepackage{stmaryrd}
\usepackage{enumitem}
\usepackage{booktabs}
\usepackage{braket}
\usepackage{multirow}
\usepackage{adjustbox}
\usepackage{pgfplots}
\usepackage{layouts}
\usepackage{tikz}
\usepackage{xcolor}
\usepackage{hhline}
\usepackage{float}
\usepackage{scrextend}
\usepackage[linesnumbered,ruled,vlined]{algorithm2e}

\usepackage[mathscr]{euscript}
\usepackage[switch]{lineno}
\usepackage{wrapfig}
\usepackage[toc,page]{appendix}

\newcommand\mytilde{\mathrel{\stackrel{\makebox[0pt]{\mbox{\normalfont\tiny i.i.d.}}}{\sim}}}
\DeclareMathOperator*{\topb}{top\_b}
\newcommand{\norm}[1]{\|#1\|}
\newcommand{\normFro}[1]{\norm{#1}_{_{F}}}

\definecolor{ForestGreen}{RGB}{34,139,34}
\definecolor{Turquoise}{RGB}{70,189,198}

\SetKwInput{KwInput}{Input}
\SetKwInput{KwOutput}{Output}
\SetKwInput{KwReturn}{Return}

\SetCommentSty{mycommfont}
\SetKwComment{Comment}{$\triangleright$\ }{}

\usepackage{amsthm}
\usepackage{thmtools}
\newtheorem{definition}{Definition}

\title{Conditionally Adaptive Multi-Task Learning: Improving Transfer Learning in NLP \\ Using Fewer Parameters \& Less Data}

\author{Jonathan Pilault$^{1}$\thanks{Joint first-authors}$\:$, Amine El hattami$^{1*}$, Christopher Pal$^{1,2,3}$ \\ 
$^1$Polytechnique Montreal \& Mila, $^2$Element AI,
$^3$Canada CIFAR AI Chair \\
\texttt{\{jonathan.pilault,amine.elhattami,christopher.pal\}@polymtl.ca}
}

\iclrfinalcopy 
\begin{document}

\maketitle

\begin{abstract}

Multi-Task Learning (MTL) networks have emerged as a promising method for transferring learned knowledge across different tasks. However, MTL must deal with challenges such as: overfitting to low resource tasks, catastrophic forgetting, and negative task transfer, or learning interference. Often, in Natural Language Processing (NLP), a separate model per task is needed to obtain the best performance. However, many fine-tuning approaches are both parameter inefficient, i.e., potentially involving one new model per task, and highly susceptible to losing knowledge acquired during pretraining. 
We propose a novel Transformer based Adapter consisting of a new conditional attention mechanism as well as a set of task-conditioned modules that facilitate weight sharing. Through this construction, we achieve more efficient parameter sharing and mitigate forgetting by keeping half of the weights of a pretrained model fixed. 
We also use a new multi-task data sampling strategy to mitigate the negative effects of data imbalance across tasks. Using this approach, we are able to surpass single task fine-tuning methods while being parameter and data efficient (using around 66\% of the data for weight updates). Compared to other BERT Large methods on GLUE, our 8-task model surpasses other Adapter methods by 2.8\% and our 24-task model outperforms by 0.7-1.0\% models that use MTL and single task fine-tuning. We show that a larger variant of our single multi-task model approach performs competitively across 26 NLP tasks and yields state-of-the-art results on a number of test and development sets. Our code is publicly available at \url{https://github.com/CAMTL/CA-MTL}.


\end{abstract}

\section{Introduction}
The introduction of deep, contextualized Masked Language Models (MLM)\footnote{For reader convenience, all acronyms in this paper are summarized in section \ref{append:acronyms} of the Appendix.} trained on massive amounts of unlabeled data has led to significant advances across many different Natural Language Processing (NLP) tasks \citep{word_context,trans_context}. Much of these recent advances can be attributed to the now well-known BERT approach \citep{bert}. Substantial improvements over previous state-of-the-art results on the GLUE benchmark
\citep{wang-etal-2018-glue} have been obtained by multiple groups using BERT models with task specific fine-tuning. The ``BERT-variant + fine-tuning'' formula has continued to improve over time with newer work constantly pushing the state-of-the-art forward on the GLUE benchmark. 
The use of a single neural architecture for multiple NLP tasks has shown promise long before the current wave of BERT inspired methods \citep{DBLP:conf/icml/CollobertW08} and recent work has argued that autoregressive language models (ARLMs) trained on large-scale datasets -- such as the GPT family of models \citep{Radford2018ImprovingLU}, are in practice multi-task learners \citep{brown2020language}. However, even with MLMs and ARLMs trained for multi-tasking, single task fine-tuning is usually also employed to achieve state-of-the-art performance on specific tasks of interest. 
Typically this fine-tuning process may entail: creating a task-specific fine-tuned model \citep{bert}, training specialized model components for task-specific predictions \citep{DBLP:journals/corr/abs-1902-00751} or fine-tuning a single multi-task architecture \citep{mtl_bert_liu2019}.

\begin{wrapfigure}[17]{r}{0.47\textwidth}
    \begin{center}
      \vspace{-23pt}
      \includegraphics[width=0.47\textwidth]{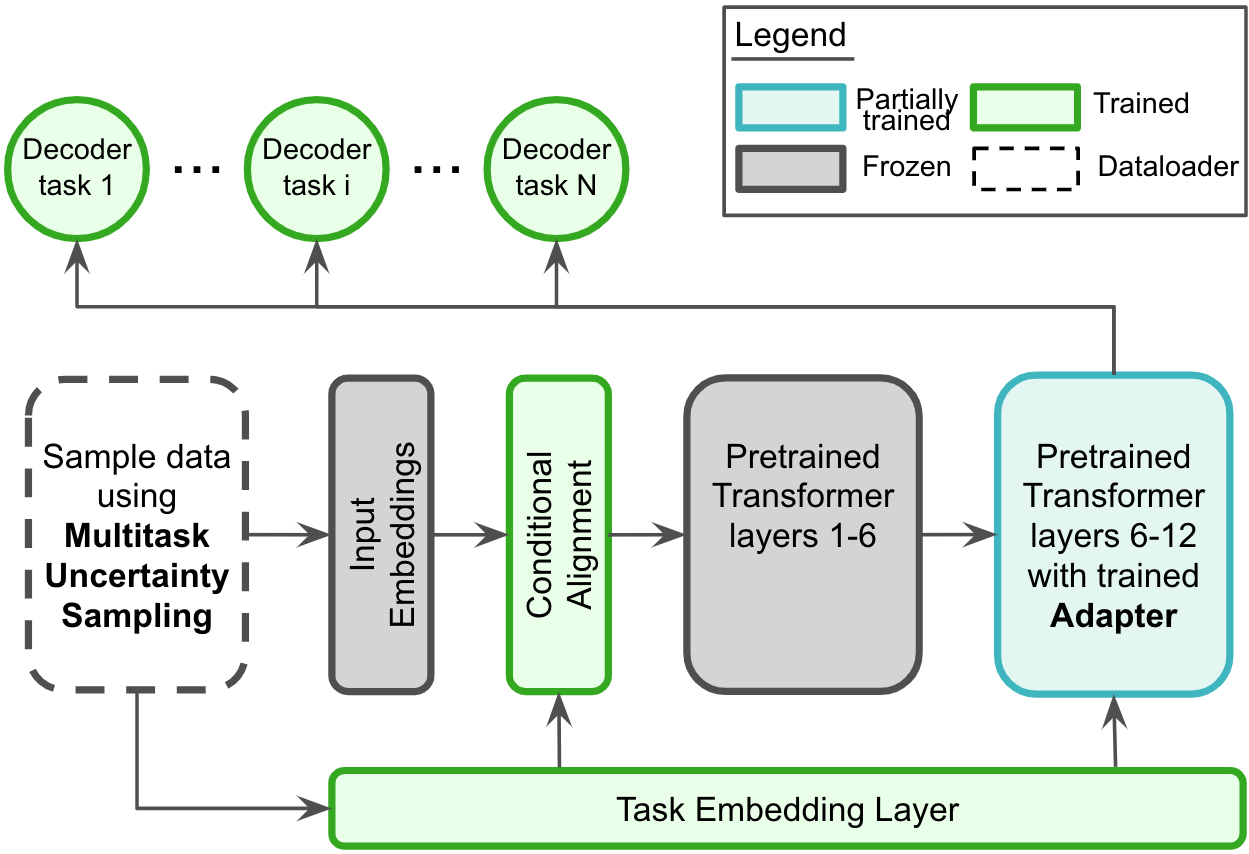}
    \end{center}
    \caption{\label{fig:model} \small $\text{CA-MTL}$ base architecture with our uncertainty-based sampling algorithm. Each task has its own decoder. The input embedding layer and the lower Transformer layers are frozen. The upper Transformer layer and Conditional Alignment module are modulated with the task embedding.}
\end{wrapfigure}

\emph{Single-task} fine-tuning overall pretrained model parameters may have other issues. Recent analyses of such MLM have shed light on the linguistic knowledge that is captured in the hidden states and attention maps \citep{bert_analysis,bert_classicnlp,merchant2020happens}. 
Particularly, BERT has middle Transformer \citep{transformer} layers that are typically the most transferable to a downstream task \citep{trans_context}. The model proxies the steps of the traditional NLP pipeline in a localizable way \citep{bert_classicnlp} 
--- with basic syntactic information appearing earlier in the network, while high-level semantic information appearing in higher-level layers. Since pretraining is usually done on large-scale datasets, it may be useful, for a variety of downstream tasks, to conserve that knowledge. However, single task fine-tuning causes catastrophic forgetting of the knowledge learned during MLM \citep{howard-ruder-2018-universal}. To preserve knowledge, freezing part of a pretrained network and using \emph{Adapters} for new tasks have shown promising results \citep{DBLP:journals/corr/abs-1902-00751}.

Inspired by the human ability to transfer learned knowledge from one task to another new task, Multi-Task Learning (MTL) in a general sense \citep{mtl_caruana1997,squad_rajpurkar2016,Ruder2017AnOO} has been applied in many fields outside of NLP. \citet{Caruana93multitasklearning} showed that a model trained in a \emph{multi-task} manner can take advantage of the inductive transfer between tasks, achieving a better generalization performance. MTL has the advantage of computational/storage efficiency \citep{DBLP:journals/corr/ZhangY17aa}, but training models in a multi-task setting is a balancing act; particularly with datasets that have different: \textbf{(a)} dataset sizes, \textbf{(b)} task difficulty levels, and \textbf{(c)} different types of loss functions. In practice, learning multiple tasks at once is challenging since negative transfer \citep{wang-acl-19}, task interference \citep{Wu2020Understanding,yu2020gradient} and catastrophic forgetting \citep{DBLP:conf/icml/SerraSMK18} can lead to worse data efficiency, training stability and generalization compared to single task fine-tuning.

Using Conditionally Adaptive Learning, we seek to improve pretraining knowledge retention and multi-task inductive knowledge transfer. Our contributions are the following: 
\begin{itemize}[noitemsep,partopsep=0pt,topsep=0pt,parsep=0pt,leftmargin=4mm]
  \item A new task conditioned Transformer that adapts and modulates pretrained weights \textbf{(Section \ref{sec:conditional_transformer})}.
  \item A novel way to prioritize tasks with an uncertainty based multi-task data sampling method that helps balance the sampling of tasks to avoid catastrophic forgetting \textbf{(Section \ref{sec:uncert_samp})}.
\end{itemize}
Our Conditionally Adaptive Multi-Task Learning (CA-MTL) approach is illustrated in Figure \ref{fig:model}. To the best of our knowledge, our work is the first to explore the use of a latent representation of tasks to modularize and adapt pretrained architectures. Further, we believe our work is also the first to examine  uncertainty sampling for large-scale multi-task learning in NLP. We show the efficacy of CA-MTL by: \textbf{(a)} testing on 26 different tasks and \textbf{(b)} presenting state-of-the-art results on a number of test sets as well as superior performance against both single-task and MTL baselines. Moreover, we further demonstrate that our method has advantages over \textbf{(c)} other adapter networks, and \textbf{(d)} other MTL sampling methods. Finally, we provide ablations and separate analysis of the MT-Uncertainty Sampling technique in section \ref{sec:mt-uncert-xp} and of each component of the adapter in \ref{sec:module-analysis}.

\section{Methodology}
\label{sec:cond_param_share}

This section is organized according to the two main MTL problems that we will tackle: (1) How to modularize a pretrained network with latent task representations? (2) How to balance different tasks in MTL?
We define each task as:
$
    \mathscr{T}_i \triangleq \{ p_i(\textbf{y}_i|\textbf{x}_i,\textbf{z}_i), \mathscr{L}_i , \tilde{p}_i(\textbf{x}_i) \}
\label{eq:task_def}
$, 
where $\textbf{z}_i$ is task $i$'s learnable shallow embedding, $\mathscr{L}_i$ is the task loss, and $\tilde{p}_i(\textbf{x}_i)$ is the empirical distribution of the training data pair $\{\textbf{x}_i, \textbf{y}_i\}$, for $i \in \{1,\dotsc,T\}$ and $T$ the number of supervised tasks.
The MTL objective is:
\begin{equation}
    \underset{\phi(\textbf{z}),\theta_1,\dotsc,\theta_T}{\min}\sum\limits_{i=1}^T\mathscr{L}_i(f_{\phi(\textbf{z}_i),\theta_i}(\textbf{x}_i), \textbf{y}_i)
\label{eq:mt_obj}
\end{equation}
where $f$ is the predictor function (includes encoder model and decoder heads), $\phi(\textbf{z})$ are learnable generated weights conditioned on $\textbf{z}$, and $\theta_i$ are task-specific parameters for the output decoder heads. \textbf{z} is constructed using an embedding lookup table.
%

\subsection{Task Conditioned Transformer}
\label{sec:conditional_transformer}
Our task conditioned Transformer architecture is based on one simple concept. 
We either add conditional layers or modulate existing pretrained weights using a task representation by extending Feature Wise Linear Modulation \citep{Perez2018FiLMVR} functions in several ways depending on the Transformer layer. 
We define our framework below.
\begin{definition}[Conditional Weight Transformations]
Given a neural network weight matrix $\textbf{W}$, we compute transformations of the form $\phi(\textbf{W}|\textbf{z}_i)=\gamma_{i}(\textbf{z}_i) \textbf{W} + \beta_{i}(\textbf{z}_i)$, where $\gamma_i$ and $\beta_i$ are learned functions that transform the weights based on a learned vector embedding $\textbf{z}_i$, for task $i$. 
\end{definition}
\begin{definition}[Conditionally Adaptive Learning]
In our setting, Conditionally Adaptive Learning is the process of learning a set of $\phi$s for the conditionally adaptive modules presented below along with a set of task embedding vectors $\textbf{z}_i$ for $T$ tasks, using a 
multi-task loss (see equation \ref{eq:mt_obj}).
\end{definition}
In the subsections that follow:
We introduce a new Transformer Attention Module using block-diagonal Conditional Attention that allows the original query-key based attention to account for task-specific biases (section \textbf{\ref{sec:cond_attn_mat}}). 
We propose a new Conditional Alignment method that aligns the data of diverse tasks and that performs better than its unconditioned and higher capacity predecessor (section \textbf{\ref{sec:alignment}}). We adapt layer normalization statistics to specific tasks using a new Conditional Layer Normalization module (section \textbf{\ref{sec:cond_layer_norm}}). We add a Conditional Bottleneck that facilitates  weight sharing and task-specific information flow from lower layers (section \textbf{\ref{sec:other_adaptive_layers}}). In our experiments we provide an ablation study of these components (Table \ref{table:architectural_ablation}) examining performance in terms of GLUE scores.

\subsubsection{Conditional Attention}
\label{sec:cond_attn_mat}

\begin{wrapfigure}[7]{r}{0.4\textwidth}
    \begin{center}
    \vspace{-45pt}
    {\includegraphics[width=0.4\textwidth]
    {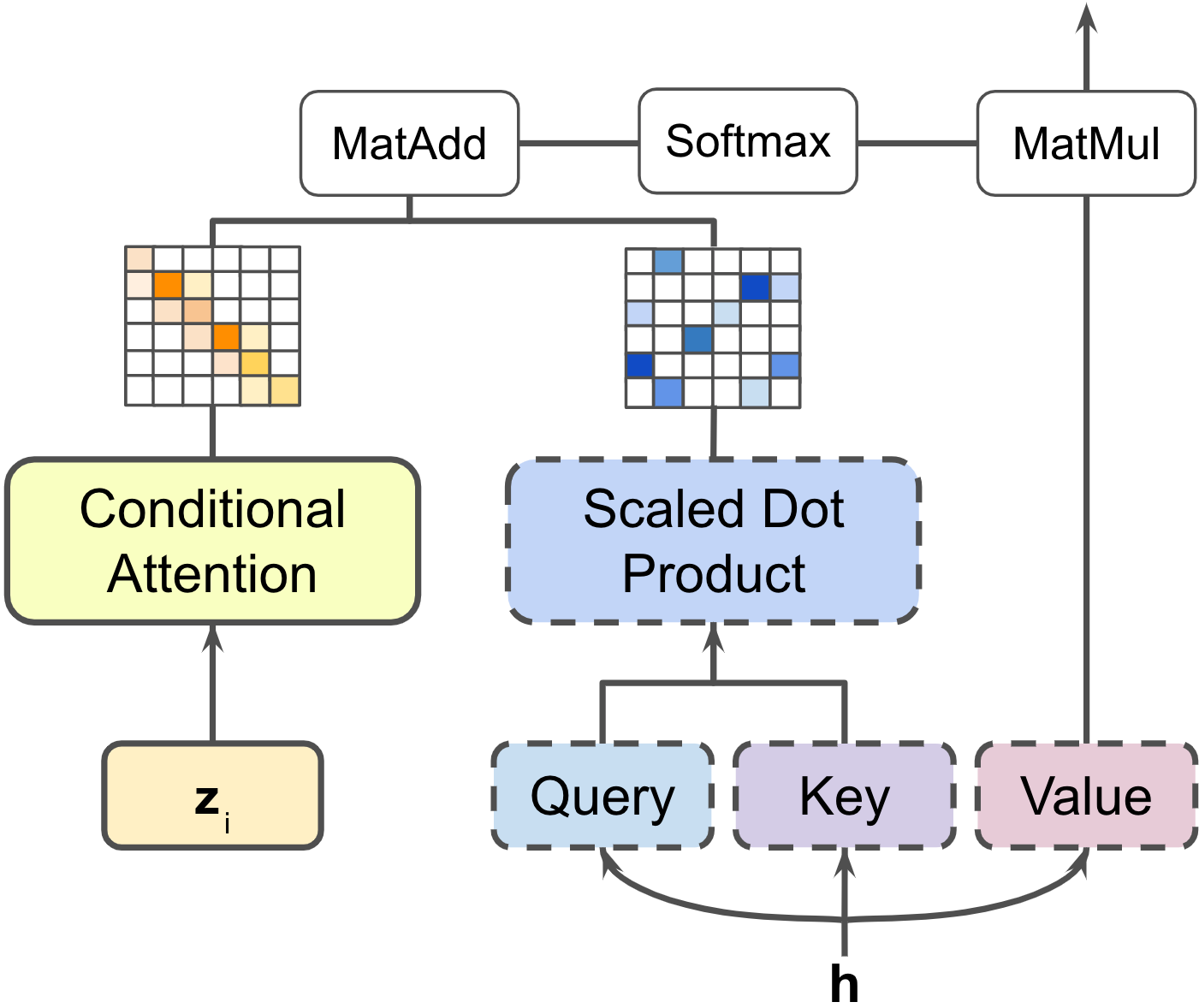}}
    \caption{\label{fig:cond_attn_mat} \small Conditional Attention Module}
    \end{center}
\end{wrapfigure}


Given $d$, the input dimensions, the query $\textbf{Q}$, the key $\textbf{K}$, and the value $\textbf{V}$ as defined in \citet{transformer}, we redefine the attention operation:
\begin{gather*}
    \textmd{Attention}(\textbf{Q}, \textbf{K}, \textbf{V}, \textbf{z}_i)) = \textmd{softmax}\Bigg[M(\textbf{z}_i) + \frac{ \textbf{Q}\textbf{K}^T}{\sqrt{d}}\Bigg] \textbf{V} \\
    M(\textbf{z}_i) = \bigoplus\limits_{n=1}^N A'_n (\textbf{z}_i), \:\:\:\: 
    A'_n(\textbf{z}_i) = A_n\gamma_{i}(\textbf{z}_i) + \beta_{i}(\textbf{z}_i)
\label{eq:cond_multihead}
\end{gather*}

where $\bigoplus$ is the direct sum operator (see section \ref{append:directsum}), $N$ is the number of block matrices $A_n \in \mathbb{R}^{(L/N)\times(L/N)}$ along the diagonal of the attention matrix, $L$ is the input sequence, $M(\textbf{z}_i)= \text{diag}(A'_1,\dotsc,A'_N)$ is a block diagonal conditional matrix. Note that $A_n$ is constructed using $L/N$ trainable and randomly initialized $L/N$ dimensional vectors. While the original attention matrix depends on the hidden states $h$, $M(\textbf{z}_i)$ is a learnable weight matrix that only depends on the task embedding $\textbf{z}_i\in \mathbb{R}^{d}$. $\gamma_{i},\beta_{i}: \mathbb{R}^{d} \mapsto \mathbb{R}^{L^2/N^2}$ are Feature Wise Linear Modulation \citep{Perez2018FiLMVR} functions. 
We also experimented with full-block Conditional Attention $\in \mathbb{R}^{L \times L}$. Not only did it have $N^2$ more parameters compared to the block-diagonal variant, but it also performed significantly worse on the GLUE development set (see FBA variant in Table \ref{table:glue_freeze}). It is possible that GLUE tasks derive a certain benefit from localized attention that is a consequence of $M(\textbf{z}_i)$.  With $M(\textbf{z}_i)$, each element in a sequence can only attend to other elements in its subsequence of length $L/N$. In our experiments we used $N=d/L$. The full Conditional Attention mechanism used in our experiments is illustrated in Figure \ref{fig:cond_attn_mat}.

\subsubsection{Conditional Alignment}
\label{sec:alignment}
\citet{Wu2020Understanding} showed that in MTL having $T$ separate alignment modules $R_1,\dotsc,R_T$ increases $\text{BERT}_{\text{LARGE}}$ avg. scores on five GLUE tasks (CoLA, MRPC, QNLI, RTE, SST-2) by 2.35\%. Inspired by this work, we found that adding a task conditioned alignment layer between the input embedding layer and the first BERT Transformer layer improved multi-task model performance. However, instead of having $T$ separate alignment matrices $R_i$ for each $T$ task, one alignment matrix $\hat{R}$ is generated as a function of the task embedding $z_i$. As in \citet{Wu2020Understanding}, we tested this module on the same five GLUE tasks and with $\text{BERT}_{\text{LARGE}}$. Enabling task conditioned weight sharing across covariance alignment modules allows us to outperforms $\text{BERT}_{\text{LARGE}}$ by 3.61\%.  This is 1.26 \% higher than having $T$ separate alignment matrices. Inserting $\hat{R}$ into BERT, yields the following encoder function $\hat{f}$:
\begin{gather}
    \hat{f} = \sum\limits_{t=1}^T g_{\theta_i}(E(\textbf{x}_i)\hat{R}(\textbf{z}_i)B), \:\:\: \:\:\: \:\:\: 
    \hat{R}(\textbf{z}_i) = R\gamma_{i}(\textbf{z}_i) + \beta_{i}(\textbf{z}_i) \label{eq:align}
\end{gather}
where $\textbf{x}_i \in \mathbb{R}^{d}$ is the layer input, $g_{\theta_i}$ is the decoder head function for task $i$ with weights $\theta_i$, $E$ the frozen BERT embedding layer, $B$ the BERT Transformer layers and $R$ the linear weight matrix of a single task conditioned alignment matrix. $\gamma_{i},\beta_{i}: \mathbb{R}^{d} \mapsto \mathbb{R}^{d}$ are Feature Wise Linear Modulation functions. 

\subsubsection{Conditional Layer Normalization (CLN)}
\label{sec:cond_layer_norm}

We extend the Conditional Batch Normalization idea from \citet{NIPS2017_7237} to Layer Normalization \citep{DBLP:journals/corr/BaKH16}. For task $\mathscr{T}_i$, $i \in \{1,\dotsc,T\}$:
\begin{gather}
\label{eqn:cond_layer_norm}
    \textbf{h}_i = \frac{1}{\sigma}\odot(\textbf{a}_i-\mu) * \hat{\gamma}_i(\textbf{z}_i) + \beta_i(\textbf{z}_i), \:\:\: \:\:\: \:\:\:
    \hat{\gamma}_i(\textbf{z}_i) = \boldsymbol{\gamma}'\gamma_i(\textbf{z}_i) + \boldsymbol{\beta}' \:\:\: \:\:\: \:\:\:
\end{gather}
where $\textbf{h}_i$ is the CLN output vector, $\textbf{a}_i$ are the preceding layer activations associated with task $i$, $\mu$ and $\sigma$ are the mean and the variance of the summed inputs within each layer as defined in \citet{DBLP:journals/corr/BaKH16}. Conditional Layer Normalization is initialized with BERT's Layer Normalization affine transformation weights and bias $\boldsymbol{\gamma}'$ and $\boldsymbol{\beta}'$ from the original formulation: $\textbf{h} = \frac{1}{\sigma}\odot(\textbf{a}-\mu) * \boldsymbol{\gamma}'+ \boldsymbol{\beta}'$. During training, the weight and bias functions of $\gamma_i(*)$ and $\beta_i(*)$ are always trained, while the original Layer Normalization weight may be kept fixed. This module was added to account for task specific rescaling of individual training cases. Layer Normalization normalizes the inputs across features. The conditioning introduced in equation \ref{sec:cond_layer_norm} allows us to modulate the normalization's output based on a task's latent representation.

\begin{wrapfigure}[13]{r}{0.40\textwidth}
    \vspace{-40pt}
    \begin{center} 
        \includegraphics[width=0.43\textwidth]{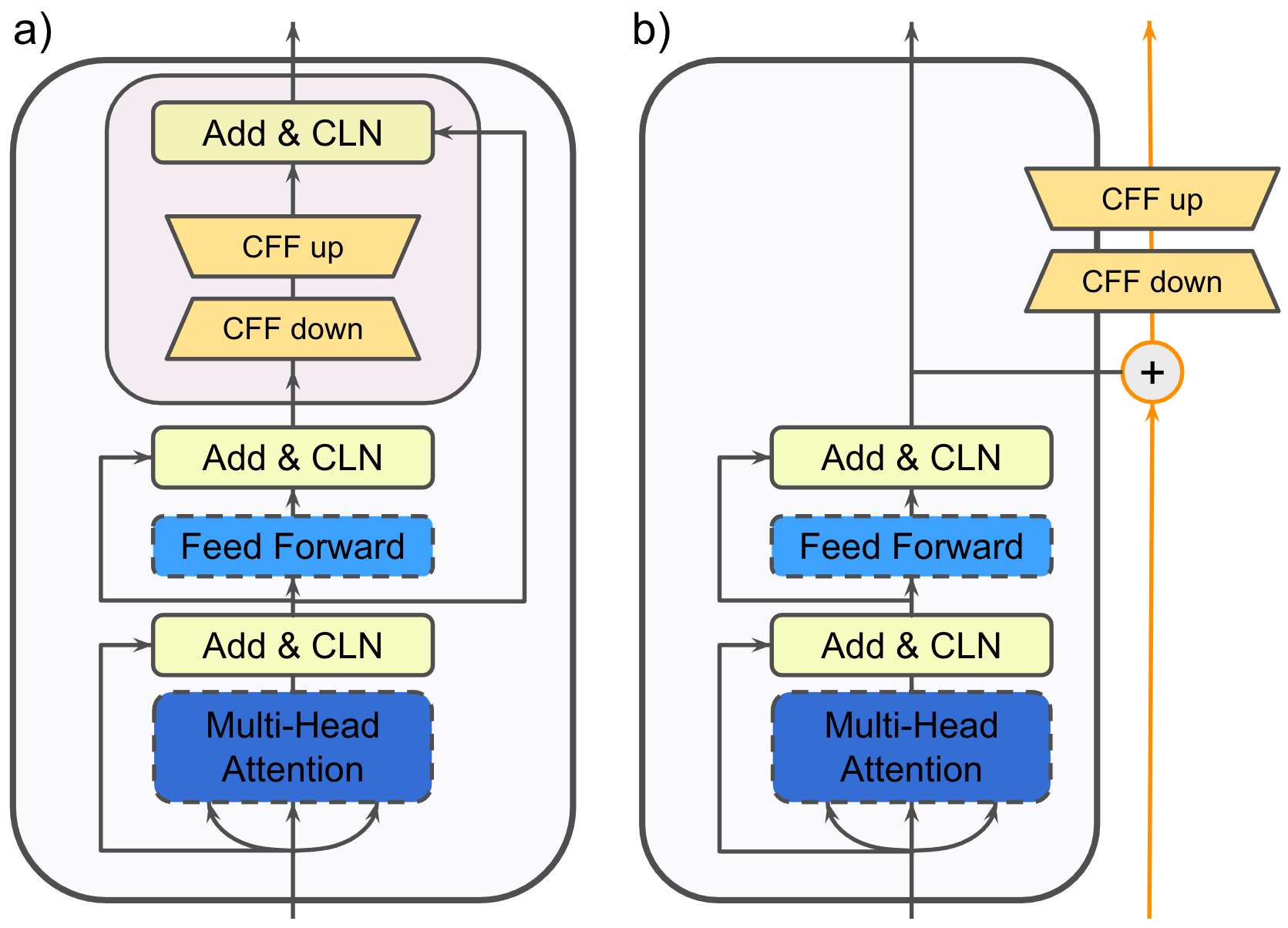}
    \caption{\label{fig:modules} \small a) Conditional Bottleneck for $\text{CA-MTL}_{\text{BASE}}$. b) Conditional Bottleneck for $\text{CA-MTL}_{\text{LARGE}}$.}
 \hspace{-5pt}
    \end{center}
\end{wrapfigure}
\subsubsection{Conditional Bottleneck}
\label{sec:other_adaptive_layers}
 We created a task conditioned two layer feed-forward bottleneck layer (CFF up/down in Figure \ref{fig:modules}). The conditional bottleneck layer follows the same transformation as in equation \ref{eq:align}. The module in Figure \ref{fig:modules}a  is added to the top most Transformer layers of $\text{CA-MTL}_{\text{BASE}}$ and uses a CLN. For $\text{CA-MTL}_{\text{LARGE}}$  this module is the main building block of the skip connection added alongside all Transformer layers seen in Figure \ref{fig:modules}b. The connection at layer $j$ takes in the matrix sum of the Transformer layer output at $j$ and the previous connection's output at $j-1$. The Conditional bottleneck allows lower layer information to flow upwards depending on the task. Our intuition for introducing this component is related to recent studies \citep{bert_classicnlp} that showed that the ``most important layers for a given task appear at specific positions''. As with the other modules described so far, each task adaptation is created from the weights of a single shared adapter that is modulated by the task embedding.

\subsection{Multi-Task Uncertainty Sampling}
\label{sec:uncert_samp}
MT-Uncertainty Sampling is a task selection strategy that is inspired by Active Learning techniques. Our algorithm \ref{alg:uncert_alg} is outlined in the Appendix, Section \ref{append:uncert_sampling}. Similar to Active Learning, our algorithm first evaluates the model uncertainty. MT-Uncertainty Sampling uses Shannon Entropy, an uncertainty measure, to choose training examples by first doing forward pass 
through the model with $b\times T$ input samples. For an output classification prediction with $C_i$ possible classes and probabilities $(p_{i,1}, \dotsc , p_{i,C_i})$, the Shannon Entropy $H_i$, for task $\mathscr{T}_i$ and $i \in \{1,\dotsc,T\}$, our uncertainty measure $\mathscr{U}(\text{x})$ are given by:
\begin{gather}
    H_i = H_i(f_{\phi(\textbf{z}_i),\theta_i}(\text{x})) = -\sum_{c=1}^{C_i} p_c\,\log\,p_c,
    \:\:\:\:\:\: \:\:\:\:\:\:
    \mathscr{U}(x_i) = \frac{H_i(f_{\phi(\textbf{z}_i),\theta_i}(\text{x}))}{\hat{H} \times H'_i}
    \label{eq:uncert}
\end{gather}
%
\begin{gather}
\vspace{-.5cm}
\hat{H} = \max_{i \in \{1,\dotsc,T\}} \bar{H}_i = \max \Bigg[\frac{1}{b} \sum_{\text{x} \in \textbf{x}_i} H_i \Bigg], \:\:\:\:\:\: \:\:\:\:\:\: 
H'_i = -\sum_{c=1}^{C_i} \frac{1}{C_i}\,\log \Bigg[\frac{1}{C_i}\Bigg]
\label{eq:norm_uni}
\end{gather}
where $\bar{H}_i$ is the average Shannon Entropy across $b$ samples of task $t$, $H'_i$, the Shannon entropy of choosing classes with uniform distribution and $\hat{H}$, the maximum of each task's average entropy over $b$ samples. $H'_i$ is normalizing factor that accounts for differing number of prediction classes (without the normalizing factor $H'_i$, tasks with a binary classification $C_i=1$ were rarely chosen). Further, to limit high entropy outliers and to favor tasks with highest uncertainty, we normalize with $\hat{H}$. The measure in eq. \ref{eq:uncert} allows Algorithm \ref{alg:uncert_alg} to choose $b$ samples from $b\times T$ candidates to train the model.

\vspace{-10pt}
\section{Related Work}
\vspace{-5pt}

\textbf{Multi-Tasking in NLP.}
To take advantage of the potential positive transfer of knowledge from one task to another, several works have proposed carefully choosing which tasks to train as an intermediate step in NLP before single task fine-tuning \citep{bingel-sogaard-2017-identifying,kerinec-etal-2018-deep,wang-acl-19,DBLP:journals/corr/abs-1905-07553,pruksachatkun2020intermediate,DBLP:journals/corr/abs-1811-01088}. The intermediate tasks are not required to perform well and are not typically evaluated jointly. In this work, all tasks are trained \emph{jointly} and \emph{all tasks used} are evaluated from a \emph{single model}. In Natural Language Understanding (NLU), it is still the case that to get the best task performance one often needs a separate model per task \citep{mtl_bert_clark2019,mccann2018natural}.
At scale, Multilingual NMT systems \citep{aharoni-etal-2019-massively} have also found that MTL model performance degrades as the number of tasks increases. We notice a similar trend in NLU with our baseline MTL model.
Recently, approaches in MTL have tackled the problem by designing task specific decoders on top of a shared model \citep{mtl_bert_liu2019} or distilling multiple single-task models into one \citep{mtl_bert_clark2019}. Nonetheless, such MTL approaches still involves single task fine-tuning. In this paper, we show that it is possible to achieve high performance in NLU without single task fine-tuning. 


\textbf{Adapters.}
Adapters are  trainable modules that are attached in specific locations of a pretrained network. They provide another promising avenue to limit the number of parameters needed when confronted with a large number of tasks. This approach is useful with pretrained MLM models that have rich linguistic information  \citep{context_prob,bert_analysis,trans_context,bert_classicnlp}.  Recently, \citet{DBLP:journals/corr/abs-1902-00751} added an adapter to a pretrained BERT model by fine-tuning the layer norms and adding feed forward bottlenecks in every Transformer layer. 
However, such methods adapt each task individually during the fine-tuning process. Unlike prior work, our method harnesses the vectorized representations of tasks to modularize a single pretrained model across all tasks. \citet{pmlr-v97-stickland19a} and \citet{tay2020hypergrid} also mix both MTL and adapters with BERT and T5 encoder-decoder \citep{t5} respectively by creating local task modules that are controlled by a global task agnostic module. The main drawback is that a new set of non-shared parameters must be added when a new task is introduced. CA-MTL shares all parameters and is able to re-modulate existing weights with a new task embedding vector.

\textbf{Active Learning, Task Selection and Sampling.}
Our sampling technique is similar to the ones found in several active learning algorithms \citep{chen-etal-2006-empirical-study} that are based on Shannon entropy estimations. \citet{reichart2008multi} and \citet{ikhwantri2018multi} examined Multi-Task Active Learning (MTAL), a technique that chooses one informative sample for $T$ different learners (or models) for each $T$ tasks. Instead we choose $T$ tasks samples for \emph{one model}.
Moreover, the algorithm weights each sample by the corresponding task score, and the Shannon entropy is normalized to account for various losses (see equation \ref{eq:norm_uni}). Also, our algorithm is used in a large scale MTL setup ($\gg$ 2 tasks). Recently, \citet{DBLP:journals/corr/abs-1907-06214}  explored task selection in MTL using learning policies based on counterfactual estimations \citep{charles2013counterfactual}. However, such method considers only fixed stochastic parameterized policies while our method \emph{adapts} its selection criterion based on model uncertainty throughout the training process.

\textbf{Hypernetworks.}
CA-MTL is a hypernetwork adapter. The method to generate \emph{task-conditioned} adapter weights is inspired by \cite{Oswald2020Continual}. Hypernetwork layers have also been finetuned along with pretrained models. For example, \cite{ponti-etal-2021-parameter} uses stochastic variational inference \cite{JMLR:v14:hoffman13a} to produce language and task latent codes that conditionally generates the weights of a BERT prediction head, a single hypernetwork linear layer shared across multiple languages and tasks.  Unlike previous methods however, CA-MTL conditionally modulates pretrained weights and biases, attention matrices, hidden representations and normalization statistics with task embeddings. Further, CA-MTL can preserve the pretraining knowledge by freezing the underlying Transformer model. Finally, we show a synergy between our hypernetwork adapter and our active task sampling technique (see section \ref{sec:uncert_samp}) that allows CA-MTL to continue surpassing fully tuned models as we scale the number of tasks (see figure \ref{fig:data_ablation}).

\section{Experiments and Results}

We show that our adapter of section \ref{sec:cond_param_share} achieve parameter efficient transfer for 26 NLP tasks. 
Our implementation of CA-MTL is based on HuggingFace \citep{huggingface}. Hyperparameters and our experimental set-up are outlined in \ref{append:more_xp_details}. To preserve the weights of the pretrained model, CA-MTL's bottom half Transformer layers are frozen in all experiments (except in section \ref{sec:new_tasks}). We also tested different layer freezing configurations and found that freezing half the layers worked best on average (see Section \ref{append:freeze_attn_block}).

\subsection{Multi-Task Uncertainty Sampling}
\label{sec:mt-uncert-xp}

\begin{wrapfigure}[15]{r}{0.45\textwidth}
    \begin{center}
        \vspace{-30pt}
        \hspace{-10pt}
        \scalebox{0.45}{\input{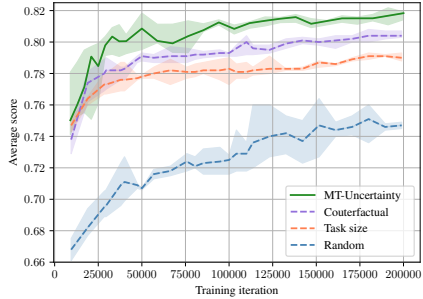}}
    \end{center}
     \caption{\label{fig:sampling_strat} \small \textcolor{ForestGreen}{\textbf{MT-Uncertainty}} vs. other task sampling strategies: median dev set scores on 8 GLUE tasks and using $\text{BERT}_{\text{BASE}}$. Data for the Counterfactual and Task Size policy $\pi_{\lvert task \rvert}$ (eq. \ref{eq:sampling}) is from \citet{DBLP:journals/corr/abs-1907-06214}.}
\end{wrapfigure}

Our MT-Uncertainty sampling strategy, from section \ref{sec:uncert_samp}, is compared to 3 other task selection schemes: a) Counterfactual b) Task size c) Random. We used a $\text{BERT}_{\text{BASE}}$ (no adapters) on 200k iterations and with the same hyperparameters as in \citet{DBLP:journals/corr/abs-1907-06214}. For more information on Counterfactual task selection, we invite the reader to consult the full explanation in \citet{DBLP:journals/corr/abs-1907-06214}. 
For $T$ tasks and the dataset $D_i$ for tasks $i \in \{1,\dotsc, T\}$, we rewrite the definitions of Random $\pi_{rand}$ and Task size $\pi_{\lvert task \rvert}$ sampling:
%
\begin{gather}
\pi_{rand} = 1/T, \:\:\: 
\pi_{\lvert task \rvert} = {\lvert D_i \rvert} \Bigg[ \sum\limits_{i=1}^T \lvert D_i \rvert\Bigg]^{-1} \label{eq:sampling}
\end{gather}
%

\begin{wrapfigure}[15]{r}{0.5\textwidth}
    \begin{center}
        \vspace{-20pt}
        \hspace*{-0.1cm}\scalebox{.55}{\input{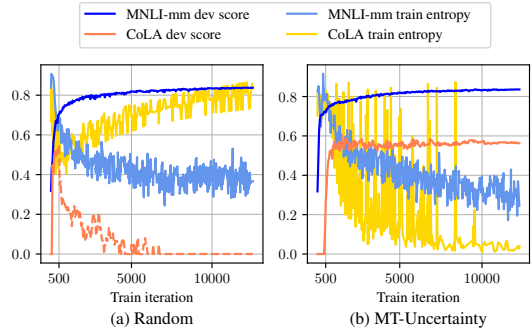}}
    \end{center}
     \caption{\label{fig:cata_forget} \small
     CoLA/MNLI Dev set scores and Entropy for $\pi_{rand}$ (left) and \textbf{MT-Uncertainty} (right).}
\end{wrapfigure}

In Figure \ref{fig:sampling_strat}, we see from the results that MT-Uncertainty converges faster by reaching the 80\% average GLUE score line before other task sampling methods. Further, MT-Uncertainty maximum score on 200k iterations is at 82.2, which is 1.7\% higher than Counterfactual sampling. The datasets in the GLUE benchmark offers a wide range of dataset sizes. This is useful to test how MT-Uncertainty manages a jointly trained low resource task (CoLA) and high resource task (MNLI). Figure \ref{fig:cata_forget} explains how catastrophic forgetting is curtailed by sampling tasks before performance drops. With $\pi_{rand}$, all of CoLA's tasks are sampled by iteration 500, at which point the larger MNLI dataset overtakes the learning process and CoLA's dev set performance starts to diminish. On the other hand, with MT-Uncertainty sampling, CoLA is sampled whenever Shannon entropy is higher than MNLI's. The model first assesses uncertain samples using Shannon Entropy then decides what data is necessary to train on. This process allows lower resource tasks to keep performance steady. We provide evidence in Figure \ref{fig:task_diff} of \ref{append:uncert_sampling} that MT-Uncertainty is able to manage task difficulty --- by choosing the most difficult tasks first.

\subsection{Ablation and Module Analysis}
\label{sec:module-analysis}

\begin{wraptable}[12]{r}{6.3cm}
\vspace{-15pt}
\caption{\small Model ablation study\textsuperscript{a} on the GLUE dev set. All models have the \textcolor{blue}{bottom half layers frozen}.}
\label{table:architectural_ablation}
\small
\setlength{\tabcolsep}{1.2pt}
\begin{tabular}{|l|c|c|c|}
	\hline 
		\multirow{2}*{Model changes} & {Avg} & {Task $\sigma$} & {\% data} \\ 
		& GLUE & GLUE & used \\
		\hline
        $\text{BERT}_\text{BASE}$ MTL ($\pi_{rand}$)  & 80.61 & 14.41 & 100 \\
        \quad + Conditional Attention   & 82.41 & 10.67 & 100 \\
        \quad + Conditional Adapter     & 82.90 & 11.27 & 100 \\
        \quad + CA and CLN              & 83.12 & 10.91 & 100 \\
        \quad + MT-Uncertainty & \multirow{2}*{\textbf{84.03}} & \multirow{2}*{\textbf{10.02}} & \multirow{2}*{66.3} \\
        \quad ($\text{CA-MTL}_\text{BERT-BASE}$)  & & & \\
    \hline
\end{tabular}
\scriptsize\textsuperscript{a}CA=Conditional Alignment, CLN=Conditional Layer Normalization, Task $\sigma$=scores standard deviation \textit{across tasks}.
\end{wraptable}
In Table \ref{table:architectural_ablation}, we present the results of an ablation study to determine which elements of $\text{CA-MTL}_{\text{BERT-BASE}}$ had the largest positive gain on average GLUE scores. Starting from a MTL $\text{BERT}_{\text{BASE}}$ baseline trained using random task sampling ($\pi_{rand}$). Apart for the Conditional Adapter, each module as well as MT-Uncertainty lift overall performance and reduce variance across tasks. Please note that we also included accuracy/F1 scores for QQP, MRPC and Pearson/ Spearman correlation for STS-B to calculate score standard deviation Task $\sigma$. Intuitively, when negative task transfer occurs between two tasks, either (1) task interference is  bidirectional and scores are both impacted, or (2) interference is unidirectional and only one score is impacted. We calculate Task $\sigma$ to characterize changes in the dynamic range of performance across multiple tasks. We do this to asses the degree to which performance improvements are distributed across all tasks or only subsets of tasks. As we can see from Table \ref{table:architectural_ablation}, Conditional Attention, Conditional Alignment, Conditional Layer Normalization, MT-Uncertainty play roles in reducing Task $\sigma$ and increasing performance across tasks. This provides partial evidence of CA-MTL's ability to mitigating negative task transfer.

\begin{wrapfigure}[14]{r}{0.45\textwidth}
    \begin{center}
        \vspace{-16pt}
        \hspace{-5pt}
        \includegraphics[width=0.46\textwidth]{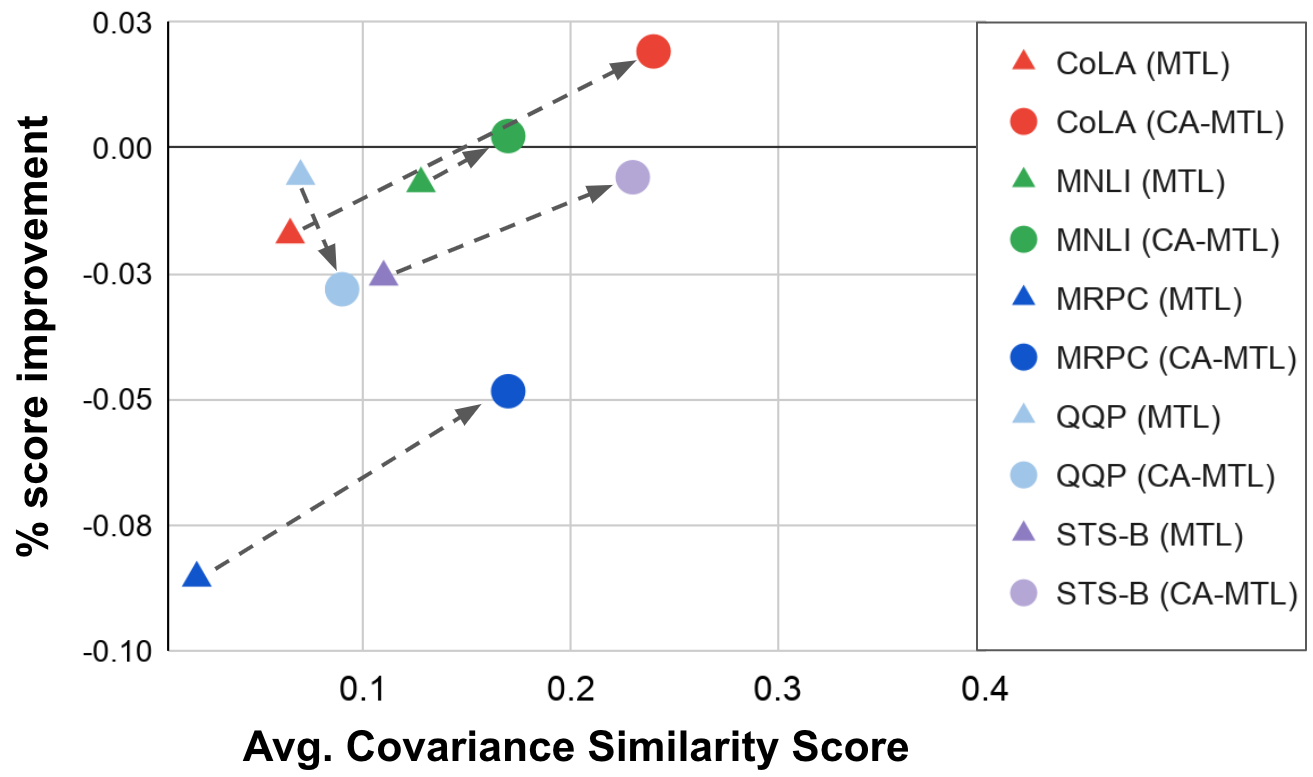}
    \end{center}
     \caption{\label{fig:covariance} \small
     Task performance vs. avg. covariance similarity scores (eq. \ref{eq_cov_sim}) for MTL and CA-MTL.}
\end{wrapfigure}

 We show that Conditional Alignment can learn to capture covariate distribution differences with task embeddings co-learned from other adapter components of CA-MTL. In Figure \ref{fig:covariance}, we arrive at similar conclusions as \cite{Wu2020Understanding}, who proved that negative task transfer is reduced when task covariances are aligned. The authors provided a ``covariance similarity score'' to gauge covariance alignment. For task $i$ and $j$ with $m_i$ and $m_j$ data samples respectively, and given $d$ dimensional inputs to the first Transformer layer $X_i \in\mathbb{R}^{m_i\times d}$ and $X_j \in\mathbb{R}^{m_j\times d}$, we rewrite the steps to calculate the covariance similarity score between task $i$ and $j$:
(a) Take the covariance matrix $X_i^{\top}X_i$,
(b) Find its best rank-$r_i$ approximation $U_{i,r_i}D_{i,r_i}U_{i,r_i}^{\top}$, where $r_i$ is chosen to contain $99\%$ of the singular values.
(c) Apply steps (a), (b) to $X_j$, and compute the covariance similarity score $CovSim_{i,j}$:
\begin{gather}
\label{eq_cov_sim}
  CovSim_{i,j} := {\small\frac{\normFro{(U_{i,r_i}D_{i,r_i}^{1/2})^{\top}U_{j,r_j}D_{j,r_j}^{1/2}}} {\normFro{U_{i,r_i}{D_{i,r_i}^{1/2}}} \cdot \normFro{U_{j,r_j}D_{j,r_j}^{1/2}}}}. \:\:\: 
  CovSim_{i} = \frac{1}{T-1}\sum\limits_{j\neq i}CovSim_{i,j}
\end{gather}

Since we are training models with $T$ tasks, we take the average covariance similarity score $CovSim_{i}$ between task $i$ and all other tasks. We measure $CovSim_{i}$ using equation \ref{eq_cov_sim} between 9 single-task models trained on individual GLUE tasks. For each task in Figure \ref{fig:covariance}, we measure the similarity score on the MTL trained $\text{BERT}_{\text{BASE}}$ baseline, e.g., CoLA (MTL), or $\text{CA-MTL}_{\text{BERT-BASE}}$ model, e.g., MNLI (CA-MTL). Our score improvement measure is the \% difference between a single task model and MTL or CA-MTL on the particular task. We find that covariance similarity increases for 9 tasks and that performance increases for 7 out 9 tasks. These measurements confirm that the Conditional Alignment is able to align task covariance, thereby helping alleviate task interference.

\subsection{Jointly training on 8 tasks: GLUE}
\label{sec:adapters}
In Table \ref{table:adapter_results}, we evaluate the performance of CA-MTL against single task fine-tuned models, MTL as well as the other BERT-based adapters on GLUE. As in \citet{DBLP:journals/corr/abs-1902-00751}, $\text{MNLI}_{\text{m}}$ and $\text{MNLI}_{\text{mm}}$ are treated as separate tasks. Our results indicate that CA-MTL outperforms both the BASE adapter, 
\begin{table*}[h!]
\vspace{-5pt}
\caption{\small \textcolor{blue}{Adapters with layer freezing} vs. ST/MT on GLUE test set. F1 scores are reported for QQP/MRPC, Spearman's correlation for STS-B, accuracy on the matched/mismatch sets for MNLI, Matthew's correlation for CoLA and accuracy for other tasks. * Individual scores not available. ST=Single Task, MTL=Multitask, g.e.= greater or equal to.
Results from:
$^{1}$\citet{bert}
$^{2}$\citet{pmlr-v97-stickland19a}.
$^{3}$\citet{DBLP:journals/corr/abs-1902-00751}
.}
\label{table:adapter_results}
\begin{center}
\scriptsize
\setlength{\tabcolsep}{2pt}
\begin{tabular}{|c|c|c|c|c|ccccccccc|c}
	\hline 
		\multirow{2}*{Method} & \multirow{2}*{Type} & Total & Trained & \# tasks
		& \multicolumn{9}{c|}{GLUE}  \\
        & & params & params/task & g.e. ST & CoLA & MNLI & MRPC & QNLI & QQP & RTE & SST-2 & STS-B & Avg \\ \hline
        
        \hline
        \multicolumn{14}{|c|}{\textbf{Base Models --- Test Server Results}} \\
        \hline
        $\text{BERT}_{\text{BASE}}$$^{1}$ & ST & 9.0$\times$ & 100\% & --- & 52.1 & 84.6/83.4 & \underline{88.9} & \underline{90.5} & 71.2 & 66.4 & \underline{93.5} & \underline{85.8} & 79.6 \\
        $\text{BERT}_{\text{BASE}}$$^{2}$ & MTL & \underline{1.0}$\times$ & 11.1\% & 2 & 51.2 & 84.0/83.4 & 86.7 & 89.3 & 70.8 & \underline{76.6} & 93.4 & 83.6 & 79.9 \\
        
        PALs+Anneal Samp.$^{2}$  & MTL & 1.13$\times$ & 12.5\% & 4 & 51.2 & 84.3/83.5 & 88.7 & 90.0 & \underline{71.5} & 76.0 & 92.6 & \underline{85.8} & 80.4 \\
     
        $\textcolor{blue}{\emph{CA-MTL}_{\text{BERT-BASE}}}$ (ours)& MTL & 1.12$\times$ & \underline{5.6} \% & \textbf{5} & \underline{53.1} & \underline{85.9}/\underline{85.8} & 88.6 & \underline{90.5} & 69.2 & 76.4 & 93.2 & 85.3  & \textbf{80.9}  \\

        \hline
        \multicolumn{14}{|c|}{\textbf{Large Models --- Test Server Results}} \\
        \hline
        
        $\text{BERT}_{\text{LARGE}}$$^{1}$ & ST & 9.0$\times$ & 100\% & --- & \underline{60.5} & \underline{86.7/85.9} & 89.3 & \underline{92.7} & \underline{72.1} & 70.1 & \underline{94.9} & 86.5 & 82.1 \\
        
        $\textcolor{blue}{\text{Adapters-256}}$$^{3}$ & ST & 1.3$\times$ & 3.6\% & 3 & 59.5 & 84.9/85.1 & \underline{89.5} & 90.7 & 71.8 & 71.5 & 94.0 & 86.9 & 80.0 \\
        
        $\textcolor{blue}{\emph{CA-MTL}_{\text{BERT-LARGE}}}$ (ours)& MTL & \underline{1.12}$\times$ & 5.6\% & 3 & 59.5 & 85.9/85.4 & 89.3 & 92.6 & 71.4 & \underline{79.0} & 94.7 & \underline{87.7}  & \textbf{82.8} \\

    \hline
\end{tabular}
\end{center}
\end{table*}
 
 PALS+Anneal Sampling \citep{pmlr-v97-stickland19a}, and the LARGE adapter,
 Adapters-256 \citep{DBLP:journals/corr/abs-1902-00751}. Against single task (ST) models, CA-MTL is 1.3\% higher than $\text{BERT}_{\text{BASE}}$, with 5 out 9 tasks equal or greater performance, and 0.7\% higher than $\text{BERT}_{\text{LARGE}}$, with 3 out 9 tasks equal or greater performance. ST models, however, need 9 models or close to $9\times$ more parameters for all 9 tasks. We noted that $\text{CA-MTL}_{\text{BERT-LARGE}}$'s average score is driven by strong RTE scores. While RTE benefits from MTL, this behavior may also be a side effect of layer freezing. In Table \ref{table:glue_freeze}, we see that CA-MTL has gains over ST on more and more tasks as we gradually unfreeze layers.

\subsection{Transfer to New Tasks}
\label{sec:new_tasks}

\begin{wraptable}[6]{r}{7.2cm}
    \vspace{-30pt}
    \caption{\small Domain adaptation results on dev. sets for \emph{BASE} models. $^{1}$\citet{mtl_bert_liu2019}, $^{2}$\citet{jiang-etal-2020-smart}}
    \label{table:domain_adapt}
    \raggedright
    \scriptsize
    \setlength{\tabcolsep}{2pt}
    \begin{tabular}{|l|c|c|c|c||c|c|c|c|}
    \hline
    \multirow{2}*{\% data used} & \multicolumn{4}{c||}{SciTail} & \multicolumn{4}{c|}{SNLI} \\
    & 0.1\% & 1\%  & 10\% & 100\%  & 0.1\% & 1\%  & 10\% & 100\%  \\
    \hline
    $\text{BERT}_{\text{BASE}}$$^{1}$ & 51.2 & 82.2 & 90.5 & 94.3 & 52.5 & 78.1  & 86.7 & 91.0\\ 
    MT-DNN$^{1}$ & 81.9 & 88.3 & 91.1 & 95.7 & 81.9 & 88.3 & 91.1 & 95.7 \\
    $\text{MT-DNN}_{\text{SMART}}$$^{2}$ & 82.3 & 88.6 & 91.3 & \textbf{96.1} & 82.7 & 86.0 & \textbf{88.7} & \textbf{91.6} \\
    $\text{CA-MTL}_{\text{BERT}}$ & \textbf{83.2} & \textbf{88.7} & \textbf{91.4} & 95.6 & \textbf{82.8} & \textbf{86.2} & 88.0 & 91.5 \\ \hline
    \end{tabular}
\end{wraptable}

In Table \ref{table:domain_adapt} we examine the ability of our method to quickly adapt to new tasks. We performed domain adaptation on SciTail \citep{Khot2018SciTaiLAT} and SNLI \citep{snli:emnlp2015} datasets, using a $\text{CA-MTL}_{\text{BASE}}$ model trained on GLUE and a new linear decoder head. We tested several pretrained and randomly initialized task embeddings in a zero-shot setting. The complete set of experiments with all task embeddings can be found in the Appendix, Section \ref{append:newtask_embed_choice}. We then selected the best task embedding for our results in Table \ref{table:domain_adapt}. STS-B and MRPC MTL-trained task embeddings performed best on SciTail and SNLI respectively. $\text{CA-MTL}_{\text{BERT-BASE}}$ has faster adaptation than $\text{MT-DNN}_{\text{SMART}}$ \citep{jiang-etal-2020-smart} as evidenced by higher performances in low-resource regimes (0.1\% and 1\% of the data). When trained on the complete dataset, $\text{CA-MTL}_{\text{BERT-BASE}}$ is on par with  $\text{MT-DNN}_{\text{SMART}}$. Unlike $\text{MT-DNN}_{\text{SMART}}$ however, we do not add context from a semantic similarity model -- $\text{MT-DNN}_{\text{SMART}}$ is built off HNN \citep{he-etal-2019-hybrid}. Nonetheless, with a larger model, CA-MTL surpasses $\text{MT-DNN}_{\text{SMART}}$ on the full SNLI and SciTail datasets in Table \ref{table:ner_scitail_snli}.

\subsection{Jointly training on 24 tasks: GLUE/Super-GLUE, MRQA and WNUT2017}
\label{sec:24task_mtl}

\begin{wrapfigure}[15]{r}{0.5\textwidth}
    \vspace{-15pt}
    \begin{center} 
        \includegraphics[width=0.5\textwidth]{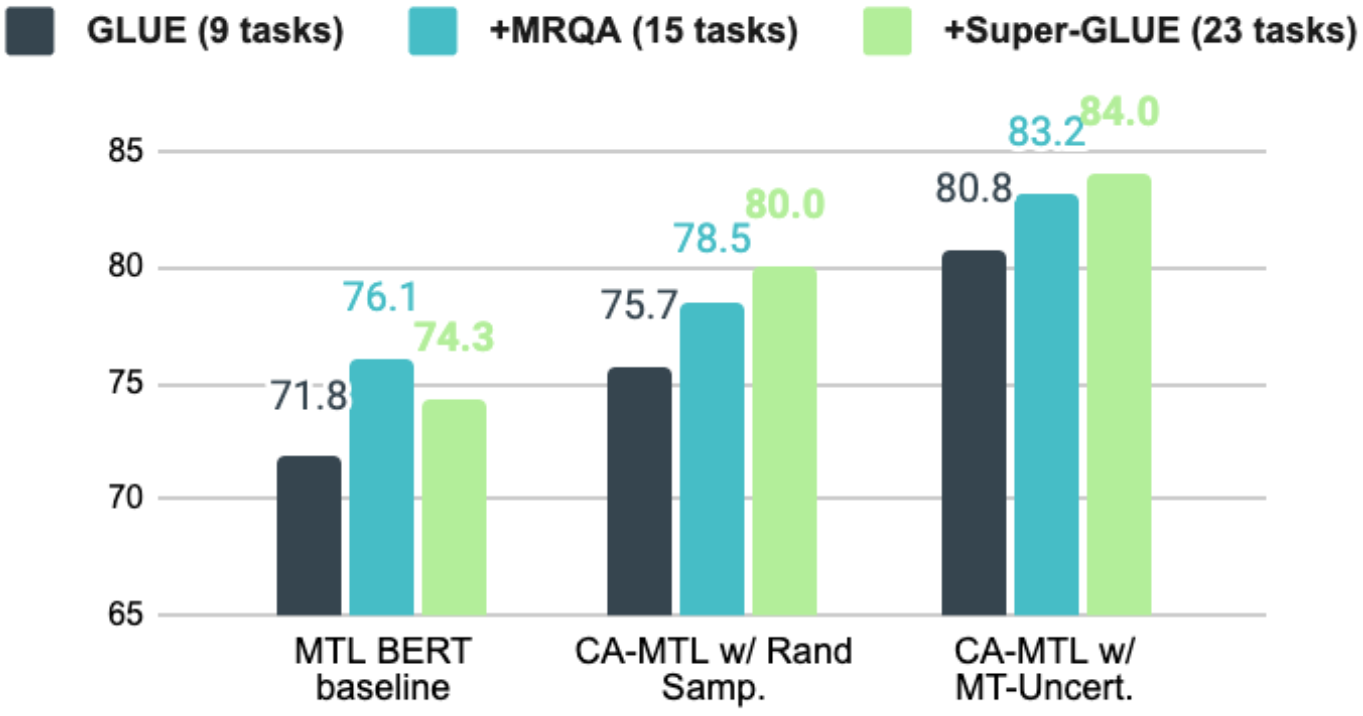}
    \caption{\label{fig:data_ablation} \small Effects of adding more datasets on avg GLUE scores. Experiments conducted on 3 epochs. When 23 tasks are trained jointly, performance of $\text{CA-MTL}_{\text{BERT-BASE}}$ continues to improve.}
    \end{center}
\end{wrapfigure}

\textbf{Effects of Scaling Task Count.} In Figure \ref{fig:data_ablation} we continue to test if CA-MTL mitigates \emph{task interference} by measuring GLUE average scores when progressively adding 9 GLUE tasks, 8 Super-GLUE tasks \citep{superglue}, 6 MRQA tasks \citep{fisch2019mrqa}. Tasks are described in Appendix section \ref{append:datasets}. The results show that adding 23 tasks drops the performance of our baseline MTL $\text{BERT}_{\text{BASE}}$ ($\pi_{rand}$). MTL BERT increases by 4.3\% when adding MRQA but, with 23 tasks, the model performance drops by 1.8\%. The opposite is true when CA-MTL modules are integrated into the model. CA-MTL continues to show gains with a large number of tasks and surpasses the baseline MTL model by close to 4\% when trained on 23 tasks.

\begin{wraptable}[13]{r}{7.2cm}
    \caption{\footnotesize 24-task CA-MTL vs. ST and vs. 24-task MTL \textcolor{blue}{with frozen 
    layers} on GLUE, SuperGLUE, MRQA and NER development sets. ST=Single Task, MTL=Multitask, g.e.= greater or equal to. Details in section \ref{append:more_xp_details}.}
    \label{table:24-task-camtl}
    \raggedright
    \scriptsize
    \setlength{\tabcolsep}{2.3pt}
    \begin{tabular}{|l|cccc|c|c|r|}
    \hline
    \multirow{2}*{Model} & \multicolumn{4}{c|}{Task Grouping} & \multirow{2}*{Avg}& \# tasks  & Total\\
                         & GLUE & SuperGLUE & MRQA & NER      &                   & e.g. ST   & Params\\
    \hline
    \multicolumn{8}{|l|}{\emph{$\text{BERT-LARGE}$ models}} \\
    \hline
    $\text{ST}_{\text{ReImp}}$      & 84.5 & 68.9 & \underline{79.7} & \underline{54.1} & 76.8 & ---    & 24$\times$\\
    $\text{MTL}_{\text{ReImp}}$     & 83.2 & 72.1 & 77.8 & 42.2 & 76.4 & 9/24   & 1$\times$\\
    \textcolor{blue}{CA-MTL}        & \underline{86.6}& \underline{74.1} & 79.5 & 49.0 & \underline{79.1} & \textbf{17/24}  & 1.12$\times$ \\
    \hline
    \multicolumn{8}{|l|}{\emph{$\text{RoBERTa-LARGE}$ models}} \\
    \hline
    $\text{ST}_{\text{ReImp}}$      & 88.2 & 76.5 & \underline{83.6} & \underline{57.8} & 81.9 & ---   & 24$\times$\\
    $\text{MTL}_{\text{ReImp}}$     & 86.0 & 78.6 & 80.7 & 49.3 & 80.7 & 7/24  & 1$\times$  \\
    \textcolor{blue}{CA-MTL}        & \underline{89.4} & \underline{80.0} & 82.4 & 55.2 & \underline{83.1} & \textbf{15/24} & 1.12$\times$\\
    \hline
    \end{tabular}
\end{wraptable}

\textbf{24-task CA-MTL.} We \emph{jointly} trained large MTL baselines and CA-MTL models on GLUE/Super-GLUE/MRQA and Named Entity Recognition (NER) WNUT2017 \citep{derczynski-etal-2017-results}. Since some dev. set scores are not provided and since RoBERTa results were reported with a median score over 5 random seeds, we ran our own single seed ST/MTL baselines (marked ``ReImp'') for a fair comparison. The dev. set numbers reported in \cite{roberta} are displayed with our baselines in Table \ref{table:glue_baselines}.  Results are presented in Table \ref{table:24-task-camtl}. We notice in Table \ref{table:24-task-camtl} that even for large models, CA-MTL provides large gains in performance on average over both ST and MTL models. For the BERT based models, CA-MTL provides 2.3\% gain over ST and higher scores on 17 out 24 tasks. For RoBERTa based models, CA-MTL provides 1.2\% gain over ST and higher scores on 15 out 24 tasks. We remind the reader that this is achieved with a single model. Even when trained with 16 other tasks, it is interesting to note that the MTL baseline perform better than the ST baseline on Super GLUE where most tasks have a small number of samples. Also, we used NER to test if we could still outperform the ST baseline on a token-level task, significantly different from other tasks. Unfortunately, while CA-MTL performs significantly better than the MTL baseline model, CA-MTL had not yet overfit on this particular task and could have closed the gap with the ST baselines with more training cycles.

\begin{wraptable}[14]{r}{8.6cm}
    \vspace{-10pt}
    \caption{\footnotesize Our 24-task CA-MTL vs. other large models on GLUE. F1 is reported for QQP/MRPC, Spearman's corr. for STS-B, Matthew's corr. for CoLA and accuracy for other tasks.
    *Split not available.
    \textcolor{red}{**}Uses intermediate task fine-tuning + ST.
    }
    \label{table:glue-sota}
    \raggedright
    \scriptsize
    \setlength{\tabcolsep}{2pt}
    \begin{tabular}{|l|cccccccc|c|}
    \hline
    \multirow{2}*{Model}& \multicolumn{8}{c|}{\textbf{GLUE tasks}} & \multirow{2}*{Avg}\\
                        & CoLA & MNLI & MRPC & QNLI & QQP & RTE & SST-2 & STS-B &  \\ \hline
    
    \hline
    \multicolumn{10}{|l|}{\emph{$\text{BERT-LARGE}$ based models on Dev set.}} \\
    \hline
    MT-DNN            & 63.5 & \underline{87.1}/\underline{86.7} & 91.0 & 92.9 & \underline{89.2} & 83.4 & 94.3 & 90.6 & 85.6 \\
    STILTS \textcolor{red}{**}           & 62.1 & 86.1*     & 92.3 & 90.5 & 88.5 & 83.4 & 93.2 & \underline{90.8} & 85.9 \\
    BAM!              & 61.8 & 87.0* & -- & 92.5 & -- & 82.8 & 93.6 & 89.7 & -- \\
    \textcolor{blue}{24-task CA-MTL}& \underline{63.8} & 86.3/86.0 & \underline{92.9} & \underline{93.4} & 88.1	& \underline{84.5} & \underline{94.5} & 90.3 & \textbf{86.6} \\
    
    \hline
    \multicolumn{10}{|l|}{\emph{$\text{RoBERTa-LARGE}$ based models on Test set.}} \\
    \hline
     RoBERTA\textcolor{red}{**} with & \multirow{2}*{\underline{67.8}} & \multirow{2}*{\underline{91.0}/\underline{90.8}} & \multirow{2}*{\underline{91.6}} & \multirow{2}*{\underline{95.4}} & \multirow{2}*{\underline{74.0}} & \multirow{2}*{\underline{87.9}} & \multirow{2}*{\underline{97.5}} & \multirow{2}*{\underline{92.5}} & \multirow{2}*{\textbf{87.3}} \\
     Ensemble             & & & & & & & & & \\
    \textcolor{blue}{24-task CA-MTL}& 62.2 & 89.0/88.4 & 92.0 & 94.7 & 72.3 & 86.2 & 96.3 & 89.8 & 85.7 \\
    \hline
    \end{tabular}
\end{wraptable}

\textbf{Comparisons with other methods.} In Table \ref{table:glue-sota}, $\text{CA-MTL}_{\text{BERT}}$ is compared to other Large BERT based methods that either use MTL + ST, such as MT-DNN \citep{mtl_bert_liu2019}, intermediate tasks + ST, such as STILTS \citep{DBLP:journals/corr/abs-1811-01088} or MTL model distillation + ST, such as BAM! \citep{mtl_bert_clark2019}. Our method scores higher than MT-DNN on 5 of 9 tasks and by 1.0 \% on avg. Against STILTS, CA-MTL realizes a 0.7 \% avg. score gain, surpassing scores on 6 of 9 tasks. We also show that $\text{CA-MTL}_{\text{RoBERTa}}$ is within only 1.6 \% of a RoBERTa ensemble of 5 to 7 models per task and that uses intermediate tasks. Using our 24-task CA-MTL large RoBERTa-based model, we report NER F1 scores on the WNUT2017 test set in Table \ref{table:ner_scitail_snli}a. 
We compare our result with $\text{RoBERTa}_{\text{LARGE}}$ and  $\text{XLM-R}_{\text{LARGE}}$ \citep{nguyen2020bertweet} the current state-of-the-art (SOTA). Our model outperforms $\text{XLM-R}_{\text{LARGE}}$ by 1.6\%, reaching a new state-of-the-art. Using domain adaptation as described in Section \ref{sec:new_tasks}, we report results on the SciTail test set in Table \ref{table:ner_scitail_snli}b and SNLI test set in Table \ref{table:ner_scitail_snli}b. For SciTail, our model matches the current SOTA\footnote{https://leaderboard.allenai.org/scitail/submissions/public on 09/27/2020} ALUM \citep{liu2020adversarial}, a RoBERTa large based model that additionally uses the SMART \citep{jiang-etal-2020-smart} fine-tuning method. For SNLI, our model outperforms SemBert, the current SOTA\footnote{https://nlp.stanford.edu/projects/snli/ on 09/27/2020}. 



\begin{table}[h]
    \caption{\small CA-MTL test performance vs. SOTA.}
    \begin{subtable}[h]{0.3\textwidth}
        \centering
        \footnotesize
        \setlength{\tabcolsep}{2pt}
        \begin{tabular}{|l|c|}
        	\hline 
        		\textcolor{teal}{\textbf{(a)}} \textbf{WNUT2017} & F1 \\
        		\hline
                $\text{RoBERTa}_{\text{LARGE}}$         & 56.9 \\
                $\text{XLM-R}_{\text{LARGE}}$           & 57.1 \\
                \hline
                $\text{CA-MTL}_{\text{RoBERTa}}$ (ours)  & \textbf{58.0} \\
            \hline
        \end{tabular}
    \end{subtable}
    \hfill
    \begin{subtable}[h]{0.3\textwidth}
        \centering
        \footnotesize
        \setlength{\tabcolsep}{2pt}
        \begin{tabular}{|l|c|}
    	\hline
    	    \textcolor{teal}{\textbf{(b)}} \textbf{SciTail} & \% Acc \\
    	    \hline
             MT-DNN                                & 94.1           \\
            $\text{ALUM}_{\text{RoBERTa}}$        & 96.3           \\
            $\text{ALUM}_{\text{RoBERTa-SMART}}$  &\textbf{96.8}   \\
        \hline
            $\text{CA-MTL}_{\text{RoBERTa}}$ (ours)     &\textbf{96.8}   \\
        \hline
    \end{tabular}

     \end{subtable}
     \hfill
    \begin{subtable}[h]{0.3\textwidth}
        \centering
        \footnotesize
        \setlength{\tabcolsep}{2pt}
        \begin{tabular}{|l|c|}
    	\hline 
        	\textcolor{teal}{\textbf{(c)}} \textbf{SNLI} & \% Acc \\
        	\hline
    		MT-DNN                                & 91.6 \\
            $\text{MT-DNN}_{\text{SMART}}$        & 91.7 \\
            SemBERT                               & 91.9 \\
            \hline
            $\text{CA-MTL}_{\text{RoBERTa}}$ (ours)     & \textbf{92.1} \\
        \hline
    \end{tabular}
     \end{subtable}
     
     \label{table:ner_scitail_snli}
\end{table}

\vspace{-.15cm}
\section{Conclusion}
We believe that our experiments here have helped demonstrate the potential of task conditioned adaptive learning within a single model that performs multiple tasks.
In a large-scale 24-task NLP experiment, CA-MTL outperforms fully tuned single task models by 2.3\% for BERT Large and by 1.2\% for RoBERTa Large using 1.12 times the number of parameters, while single task fine-tuning approach requires 24 separately tuned single task models or 24 times the number of parameters. When a BERT vanilla MTL model sees its performance drop as the number of tasks increases, CA-MTL scores continue to climb. Performance gains are not driven by a single task as it is often the case in MTL. Each CA-MTL module that adapts a Transformer model is able to reduce performance variances between tasks, increasing average scores and aligning task covariances. This evidence shows that CA-MTL is able to mitigate task interference and promote more efficient parameter sharing. We showed that MT-Uncertainty is able to avoid degrading performances of low resource tasks. Tasks are sampled whenever the model sees entropy increase, helping avoid catastrophic forgetting. Overall, CA-MTL offers a promising avenue to dynamically adapt and modularize knowledge embedded in large monolithic pretrained models. Extending such ideas will be an objective for future work.







\clearpage

\subsubsection*{Acknowledgments}
This research was supported by the Canada CIFAR AI Chairs Program, NSERC and PROMPT. Experiments in this article were conducted with Compute Canada and MILA computational infrastructure and we thank them for their support. We would like to thank Colin Raffel, Sandeep Subramanian, and Nicolas Gontier for their useful feedback and the anonymous reviewers for helpful comments, discussions and suggestions.

\bibliography{references}

\begin{thebibliography}{85}
\providecommand{\natexlab}[1]{#1}
\providecommand{\url}[1]{\texttt{#1}}
\expandafter\ifx\csname urlstyle\endcsname\relax
  \providecommand{\doi}[1]{doi: #1}\else
  \providecommand{\doi}{doi: \begingroup \urlstyle{rm}\Url}\fi

\bibitem[Aharoni et~al.(2019)Aharoni, Johnson, and
  Firat]{aharoni-etal-2019-massively}
Roee Aharoni, Melvin Johnson, and Orhan Firat.
\newblock Massively multilingual neural machine translation.
\newblock In \emph{Proceedings of the 2019 Conference of the North {A}merican
  Chapter of the Association for Computational Linguistics: Human Language
  Technologies, Volume 1 (Long and Short Papers)}, pp.\  3874--3884,
  Minneapolis, Minnesota, June 2019. Association for Computational Linguistics.
\newblock \doi{10.18653/v1/N19-1388}.
\newblock URL \url{https://www.aclweb.org/anthology/N19-1388}.

\bibitem[Ba et~al.(2016)Ba, Kiros, and Hinton]{DBLP:journals/corr/BaKH16}
Lei~Jimmy Ba, Jamie~Ryan Kiros, and Geoffrey~E. Hinton.
\newblock Layer normalization.
\newblock \emph{CoRR}, abs/1607.06450, 2016.
\newblock URL \url{http://arxiv.org/abs/1607.06450}.

\bibitem[Bengio et~al.(2009)Bengio, Louradour, Collobert, and
  Weston]{bengio2009curriculum}
Yoshua Bengio, J{\'e}r{\^o}me Louradour, Ronan Collobert, and Jason Weston.
\newblock Curriculum learning.
\newblock In \emph{Proceedings of the 26th annual international conference on
  machine learning}, pp.\  41--48, 2009.

\bibitem[Bingel \& S{\o}gaard(2017)Bingel and
  S{\o}gaard]{bingel-sogaard-2017-identifying}
Joachim Bingel and Anders S{\o}gaard.
\newblock Identifying beneficial task relations for multi-task learning in deep
  neural networks.
\newblock In \emph{Proceedings of the 15th Conference of the {E}uropean Chapter
  of the Association for Computational Linguistics: Volume 2, Short Papers},
  pp.\  164--169, Valencia, Spain, April 2017. Association for Computational
  Linguistics.
\newblock URL \url{https://www.aclweb.org/anthology/E17-2026}.

\bibitem[Bowman et~al.(2015)Bowman, Angeli, Potts, and Manning]{snli:emnlp2015}
Samuel~R. Bowman, Gabor Angeli, Christopher Potts, and Christopher~D. Manning.
\newblock A large annotated corpus for learning natural language inference.
\newblock In \emph{Proceedings of the 2015 Conference on Empirical Methods in
  Natural Language Processing (EMNLP)}. Association for Computational
  Linguistics, 2015.

\bibitem[Brown et~al.(2020)Brown, Mann, Ryder, Subbiah, Kaplan, Dhariwal,
  Neelakantan, Shyam, Sastry, Askell, et~al.]{brown2020language}
Tom~B Brown, Benjamin Mann, Nick Ryder, Melanie Subbiah, Jared Kaplan, Prafulla
  Dhariwal, Arvind Neelakantan, Pranav Shyam, Girish Sastry, Amanda Askell,
  et~al.
\newblock Language models are few-shot learners.
\newblock \emph{arXiv}, pp.\  arXiv--2005, 2020.

\bibitem[Caruana(1997)]{mtl_caruana1997}
Rich Caruana.
\newblock Multitask learning.
\newblock \emph{Mach. Learn.}, 28\penalty0 (1):\penalty0 41--75, July 1997.
\newblock ISSN 0885-6125.
\newblock \doi{10.1023/A:1007379606734}.
\newblock URL \url{https://doi.org/10.1023/A:1007379606734}.

\bibitem[Caruana(1993)]{Caruana93multitasklearning}
Richard Caruana.
\newblock Multitask learning: A knowledge-based source of inductive bias.
\newblock In \emph{Proceedings of the Tenth International Conference on Machine
  Learning}, pp.\  41--48. Morgan Kaufmann, 1993.

\bibitem[Cer et~al.(2017)Cer, Diab, Agirre, Lopez-Gazpio, and Specia]{sts-b}
Daniel Cer, Mona Diab, Eneko Agirre, I{\~n}igo Lopez-Gazpio, and Lucia Specia.
\newblock {S}em{E}val-2017 task 1: Semantic textual similarity multilingual and
  crosslingual focused evaluation.
\newblock In \emph{Proceedings of the 11th International Workshop on Semantic
  Evaluation ({S}em{E}val-2017)}, pp.\  1--14, Vancouver, Canada, August 2017.
  Association for Computational Linguistics.
\newblock \doi{10.18653/v1/S17-2001}.
\newblock URL \url{https://www.aclweb.org/anthology/S17-2001}.

\bibitem[Charles et~al.(2013)Charles, Chickering, and
  Simard]{charles2013counterfactual}
Denis Charles, Max Chickering, and Patrice Simard.
\newblock Counterfactual reasoning and learning systems: The example of
  computational advertising.
\newblock \emph{Journal of Machine Learning Research}, 14:\penalty0 3207--3260,
  November 2013.

\bibitem[Chen et~al.(2006)Chen, Schein, Ungar, and
  Palmer]{chen-etal-2006-empirical-study}
Jinying Chen, Andrew Schein, Lyle Ungar, and Martha Palmer.
\newblock An empirical study of the behavior of active learning for word sense
  disambiguation.
\newblock In \emph{Proceedings of the Human Language Technology Conference of
  the {NAACL}, Main Conference}, pp.\  120--127, New York City, USA, June 2006.
  Association for Computational Linguistics.
\newblock URL \url{https://www.aclweb.org/anthology/N06-1016}.

\bibitem[Chen et~al.(2017)Chen, Badrinarayanan, Lee, and
  Rabinovich]{DBLP:journals/corr/abs-1711-02257}
Zhao Chen, Vijay Badrinarayanan, Chen{-}Yu Lee, and Andrew Rabinovich.
\newblock Gradnorm: Gradient normalization for adaptive loss balancing in deep
  multitask networks.
\newblock \emph{CoRR}, abs/1711.02257, 2017.
\newblock URL \url{http://arxiv.org/abs/1711.02257}.

\bibitem[Clark et~al.(2019{\natexlab{a}})Clark, Lee, Chang, Kwiatkowski,
  Collins, and Toutanova]{boolq}
Christopher Clark, Kenton Lee, Ming-Wei Chang, Tom Kwiatkowski, Michael
  Collins, and Kristina Toutanova.
\newblock {B}ool{Q}: Exploring the surprising difficulty of natural yes/no
  questions.
\newblock In \emph{Proceedings of the 2019 Conference of the North {A}merican
  Chapter of the Association for Computational Linguistics: Human Language
  Technologies, Volume 1 (Long and Short Papers)}, pp.\  2924--2936,
  Minneapolis, Minnesota, June 2019{\natexlab{a}}. Association for
  Computational Linguistics.
\newblock \doi{10.18653/v1/N19-1300}.
\newblock URL \url{https://www.aclweb.org/anthology/N19-1300}.

\bibitem[Clark et~al.(2019{\natexlab{b}})Clark, Khandelwal, Levy, and
  Manning]{bert_analysis}
Kevin Clark, Urvashi Khandelwal, Omer Levy, and Christopher~D. Manning.
\newblock What does {BERT} look at? an analysis of {BERT}{'}s attention.
\newblock In \emph{Proceedings of the 2019 ACL Workshop BlackboxNLP: Analyzing
  and Interpreting Neural Networks for NLP}, pp.\  276--286, Florence, Italy,
  August 2019{\natexlab{b}}. Association for Computational Linguistics.
\newblock \doi{10.18653/v1/W19-4828}.
\newblock URL \url{https://www.aclweb.org/anthology/W19-4828}.

\bibitem[Clark et~al.(2019{\natexlab{c}})Clark, Luong, Khandelwal, Manning, and
  Le]{mtl_bert_clark2019}
Kevin Clark, Minh{-}Thang Luong, Urvashi Khandelwal, Christopher~D. Manning,
  and Quoc~V. Le.
\newblock Bam! born-again multi-task networks for natural language
  understanding.
\newblock \emph{CoRR}, abs/1907.04829, 2019{\natexlab{c}}.
\newblock URL \url{http://arxiv.org/abs/1907.04829}.

\bibitem[Collins et~al.(2018)Collins, Rozanov, and
  Zhang]{collins-etal-2018-evolutionary}
Edward Collins, Nikolai Rozanov, and Bingbing Zhang.
\newblock Evolutionary data measures: Understanding the difficulty of text
  classification tasks.
\newblock In \emph{Proceedings of the 22nd Conference on Computational Natural
  Language Learning}, pp.\  380--391, Brussels, Belgium, October 2018.
  Association for Computational Linguistics.
\newblock \doi{10.18653/v1/K18-1037}.
\newblock URL \url{https://www.aclweb.org/anthology/K18-1037}.

\bibitem[Collobert \& Weston(2008)Collobert and
  Weston]{DBLP:conf/icml/CollobertW08}
Ronan Collobert and Jason Weston.
\newblock A unified architecture for natural language processing: deep neural
  networks with multitask learning.
\newblock In \emph{ICML}, pp.\  160--167, 2008.
\newblock URL \url{https://doi.org/10.1145/1390156.1390177}.

\bibitem[de~Marneffe et~al.(2019)de~Marneffe, Simons, and Tonhauser]{cb}
Marie-Catherine de~Marneffe, Mandy Simons, and Judith Tonhauser.
\newblock The commitmentbank: Investigating projection in naturally occurring
  discourse.
\newblock \emph{Proceedings of Sinn und Bedeutung}, 23\penalty0 (2):\penalty0
  107--124, Jul. 2019.
\newblock \doi{10.18148/sub/2019.v23i2.601}.
\newblock URL
  \url{https://ojs.ub.uni-konstanz.de/sub/index.php/sub/article/view/601}.

\bibitem[de~Vries et~al.(2017)de~Vries, Strub, Mary, Larochelle, Pietquin, and
  Courville]{NIPS2017_7237}
Harm de~Vries, Florian Strub, Jeremie Mary, Hugo Larochelle, Olivier Pietquin,
  and Aaron~C Courville.
\newblock Modulating early visual processing by language.
\newblock In I.~Guyon, U.~V. Luxburg, S.~Bengio, H.~Wallach, R.~Fergus,
  S.~Vishwanathan, and R.~Garnett (eds.), \emph{Advances in Neural Information
  Processing Systems 30}, pp.\  6594--6604. Curran Associates, Inc., 2017.
\newblock URL
  \url{http://papers.nips.cc/paper/7237-modulating-early-visual-processing-by-language.pdf}.

\bibitem[Derczynski et~al.(2017)Derczynski, Nichols, van Erp, and
  Limsopatham]{derczynski-etal-2017-results}
Leon Derczynski, Eric Nichols, Marieke van Erp, and Nut Limsopatham.
\newblock Results of the {WNUT}2017 shared task on novel and emerging entity
  recognition.
\newblock In \emph{Proceedings of the 3rd Workshop on Noisy User-generated
  Text}, pp.\  140--147, Copenhagen, Denmark, September 2017. Association for
  Computational Linguistics.
\newblock \doi{10.18653/v1/W17-4418}.
\newblock URL \url{https://www.aclweb.org/anthology/W17-4418}.

\bibitem[Devlin et~al.(2018)Devlin, Chang, Lee, and Toutanova]{bert}
Jacob Devlin, Ming{-}Wei Chang, Kenton Lee, and Kristina Toutanova.
\newblock {BERT:} pre-training of deep bidirectional transformers for language
  understanding.
\newblock \emph{CoRR}, abs/1810.04805, 2018.
\newblock URL \url{http://arxiv.org/abs/1810.04805}.

\bibitem[Dolan \& Brockett(2005)Dolan and Brockett]{mrpc}
William~B. Dolan and Chris Brockett.
\newblock Automatically constructing a corpus of sentential paraphrases.
\newblock In \emph{Proceedings of the Third International Workshop on
  Paraphrasing ({IWP}2005)}, 2005.
\newblock URL \url{https://www.aclweb.org/anthology/I05-5002}.

\bibitem[Dunn et~al.(2017)Dunn, Sagun, Higgins, G{\"{u}}ney, Cirik, and
  Cho]{searchqa}
Matthew Dunn, Levent Sagun, Mike Higgins, V.~Ugur G{\"{u}}ney, Volkan Cirik,
  and Kyunghyun Cho.
\newblock Searchqa: {A} new q{\&}a dataset augmented with context from a search
  engine.
\newblock \emph{CoRR}, abs/1704.05179, 2017.
\newblock URL \url{http://arxiv.org/abs/1704.05179}.

\bibitem[Fisch et~al.(2019)Fisch, Talmor, Jia, Seo, Choi, and
  Chen]{fisch2019mrqa}
Adam Fisch, Alon Talmor, Robin Jia, Minjoon Seo, Eunsol Choi, and Danqi Chen.
\newblock {MRQA} 2019 shared task: Evaluating generalization in reading
  comprehension.
\newblock In \emph{Proceedings of the 2nd Workshop on Machine Reading for
  Question Answering}, pp.\  1--13, Hong Kong, China, November 2019.
  Association for Computational Linguistics.
\newblock \doi{10.18653/v1/D19-5801}.
\newblock URL \url{https://www.aclweb.org/anthology/D19-5801}.

\bibitem[Glover \& Hokamp(2019)Glover and
  Hokamp]{DBLP:journals/corr/abs-1907-06214}
John Glover and Chris Hokamp.
\newblock Task selection policies for multitask learning.
\newblock \emph{CoRR}, 2019.
\newblock URL \url{http://arxiv.org/abs/1907.06214}.

\bibitem[Gordon et~al.(2012)Gordon, Kozareva, and Roemmele]{copa}
Andrew Gordon, Zornitsa Kozareva, and Melissa Roemmele.
\newblock {S}em{E}val-2012 task 7: Choice of plausible alternatives: An
  evaluation of commonsense causal reasoning.
\newblock In \emph{*{SEM} 2012: The First Joint Conference on Lexical and
  Computational Semantics {--} Volume 1: Proceedings of the main conference and
  the shared task, and Volume 2: Proceedings of the Sixth International
  Workshop on Semantic Evaluation ({S}em{E}val 2012)}, pp.\  394--398,
  Montr{\'e}al, Canada, 7-8 June 2012. Association for Computational
  Linguistics.
\newblock URL \url{https://www.aclweb.org/anthology/S12-1052}.

\bibitem[Guo et~al.(2018)Guo, Haque, Huang, Yeung, and Fei-Fei]{Guo_2018_ECCV}
Michelle Guo, Albert Haque, De-An Huang, Serena Yeung, and Li~Fei-Fei.
\newblock Dynamic task prioritization for multitask learning.
\newblock In \emph{Proceedings of the European Conference on Computer Vision
  (ECCV)}, September 2018.

\bibitem[He et~al.(2019)He, Liu, Chen, and Gao]{he-etal-2019-hybrid}
Pengcheng He, Xiaodong Liu, Weizhu Chen, and Jianfeng Gao.
\newblock A hybrid neural network model for commonsense reasoning.
\newblock In \emph{Proceedings of the First Workshop on Commonsense Inference
  in Natural Language Processing}, pp.\  13--21, Hong Kong, China, November
  2019. Association for Computational Linguistics.
\newblock \doi{10.18653/v1/D19-6002}.
\newblock URL \url{https://www.aclweb.org/anthology/D19-6002}.

\bibitem[Hoffman et~al.(2013)Hoffman, Blei, Wang, and
  Paisley]{JMLR:v14:hoffman13a}
Matthew~D. Hoffman, David~M. Blei, Chong Wang, and John Paisley.
\newblock Stochastic variational inference.
\newblock \emph{Journal of Machine Learning Research}, 14\penalty0
  (4):\penalty0 1303--1347, 2013.
\newblock URL \url{http://jmlr.org/papers/v14/hoffman13a.html}.

\bibitem[Houlsby et~al.(2019)Houlsby, Giurgiu, Jastrzebski, Morrone,
  de~Laroussilhe, Gesmundo, Attariyan, and
  Gelly]{DBLP:journals/corr/abs-1902-00751}
Neil Houlsby, Andrei Giurgiu, Stanislaw Jastrzebski, Bruna Morrone, Quentin
  de~Laroussilhe, Andrea Gesmundo, Mona Attariyan, and Sylvain Gelly.
\newblock Parameter-efficient transfer learning for {NLP}.
\newblock \emph{CoRR}, abs/1902.00751, 2019.
\newblock URL \url{http://arxiv.org/abs/1902.00751}.

\bibitem[Howard \& Ruder(2018)Howard and Ruder]{howard-ruder-2018-universal}
Jeremy Howard and Sebastian Ruder.
\newblock Universal language model fine-tuning for text classification.
\newblock In \emph{Proceedings of the 56th Annual Meeting of the Association
  for Computational Linguistics (Volume 1: Long Papers)}, pp.\  328--339,
  Melbourne, Australia, July 2018. Association for Computational Linguistics.
\newblock \doi{10.18653/v1/P18-1031}.
\newblock URL \url{https://www.aclweb.org/anthology/P18-1031}.

\bibitem[Ikhwantri et~al.(2018)Ikhwantri, Louvan, Kurniawan, Abisena, Rachman,
  Wicaksono, and Mahendra]{ikhwantri2018multi}
Fariz Ikhwantri, Samuel Louvan, Kemal Kurniawan, Bagas Abisena, Valdi Rachman,
  Alfan~Farizki Wicaksono, and Rahmad Mahendra.
\newblock Multi-task active learning for neural semantic role labeling on low
  resource conversational corpus.
\newblock In \emph{Proceedings of the Workshop on Deep Learning Approaches for
  Low-Resource NLP}, pp.\  43--50, 2018.

\bibitem[Jiang et~al.(2020)Jiang, He, Chen, Liu, Gao, and
  Zhao]{jiang-etal-2020-smart}
Haoming Jiang, Pengcheng He, Weizhu Chen, Xiaodong Liu, Jianfeng Gao, and Tuo
  Zhao.
\newblock {SMART}: Robust and efficient fine-tuning for pre-trained natural
  language models through principled regularized optimization.
\newblock In \emph{Proceedings of the 58th Annual Meeting of the Association
  for Computational Linguistics}, pp.\  2177--2190, Online, July 2020.
  Association for Computational Linguistics.
\newblock \doi{10.18653/v1/2020.acl-main.197}.
\newblock URL \url{https://www.aclweb.org/anthology/2020.acl-main.197}.

\bibitem[Joshi et~al.(2017)Joshi, Choi, Weld, and Zettlemoyer]{triviaqa}
Mandar Joshi, Eunsol Choi, Daniel Weld, and Luke Zettlemoyer.
\newblock {T}rivia{QA}: A large scale distantly supervised challenge dataset
  for reading comprehension.
\newblock In \emph{Proceedings of the 55th Annual Meeting of the Association
  for Computational Linguistics (Volume 1: Long Papers)}, pp.\  1601--1611,
  Vancouver, Canada, July 2017. Association for Computational Linguistics.
\newblock \doi{10.18653/v1/P17-1147}.
\newblock URL \url{https://www.aclweb.org/anthology/P17-1147}.

\bibitem[Joshi et~al.(2019)Joshi, Chen, Liu, Weld, Zettlemoyer, and
  Levy]{DBLP:journals/corr/abs-1907-10529}
Mandar Joshi, Danqi Chen, Yinhan Liu, Daniel~S. Weld, Luke Zettlemoyer, and
  Omer Levy.
\newblock Spanbert: Improving pre-training by representing and predicting
  spans.
\newblock \emph{CoRR}, abs/1907.10529, 2019.
\newblock URL \url{http://arxiv.org/abs/1907.10529}.

\bibitem[Kendall et~al.(2017)Kendall, Gal, and
  Cipolla]{DBLP:journals/corr/KendallGC17}
Alex Kendall, Yarin Gal, and Roberto Cipolla.
\newblock Multi-task learning using uncertainty to weigh losses for scene
  geometry and semantics.
\newblock \emph{CoRR}, abs/1705.07115, 2017.
\newblock URL \url{http://arxiv.org/abs/1705.07115}.

\bibitem[Kerinec et~al.(2018)Kerinec, Braud, and
  S{\o}gaard]{kerinec-etal-2018-deep}
Emma Kerinec, Chlo{\'e} Braud, and Anders S{\o}gaard.
\newblock When does deep multi-task learning work for loosely related document
  classification tasks?
\newblock In \emph{Proceedings of the 2018 {EMNLP} Workshop {B}lackbox{NLP}:
  Analyzing and Interpreting Neural Networks for {NLP}}, pp.\  1--8, Brussels,
  Belgium, November 2018. Association for Computational Linguistics.
\newblock \doi{10.18653/v1/W18-5401}.
\newblock URL \url{https://www.aclweb.org/anthology/W18-5401}.

\bibitem[Khashabi et~al.(2018)Khashabi, Chaturvedi, Roth, Upadhyay, and
  Roth]{multirc}
Daniel Khashabi, Snigdha Chaturvedi, Michael Roth, Shyam Upadhyay, and Dan
  Roth.
\newblock Looking beyond the surface: A challenge set for reading comprehension
  over multiple sentences.
\newblock In \emph{Proceedings of the 2018 Conference of the North {A}merican
  Chapter of the Association for Computational Linguistics: Human Language
  Technologies, Volume 1 (Long Papers)}, pp.\  252--262, New Orleans,
  Louisiana, June 2018. Association for Computational Linguistics.
\newblock \doi{10.18653/v1/N18-1023}.
\newblock URL \url{https://www.aclweb.org/anthology/N18-1023}.

\bibitem[Khot et~al.(2018)Khot, Sabharwal, and Clark]{Khot2018SciTaiLAT}
Tushar Khot, A.~Sabharwal, and Peter Clark.
\newblock Scitail: A textual entailment dataset from science question
  answering.
\newblock In \emph{AAAI}, 2018.

\bibitem[Kingma \& Ba(2015)Kingma and Ba]{Kingma2015AdamAM}
Diederik~P. Kingma and Jimmy Ba.
\newblock Adam: A method for stochastic optimization.
\newblock \emph{CoRR}, abs/1412.6980, 2015.

\bibitem[Kwiatkowski et~al.(2019)Kwiatkowski, Palomaki, Redfield, Collins,
  Parikh, Alberti, Epstein, Polosukhin, Kelcey, Devlin, Lee, Toutanova, Jones,
  Chang, Dai, Uszkoreit, Le, and Petrov]{natq}
Tom Kwiatkowski, Jennimaria Palomaki, Olivia Redfield, Michael Collins, Ankur
  Parikh, Chris Alberti, Danielle Epstein, Illia Polosukhin, Matthew Kelcey,
  Jacob Devlin, Kenton Lee, Kristina~N. Toutanova, Llion Jones, Ming-Wei Chang,
  Andrew Dai, Jakob Uszkoreit, Quoc Le, and Slav Petrov.
\newblock Natural questions: a benchmark for question answering research.
\newblock \emph{Transactions of the Association of Computational Linguistics},
  2019.

\bibitem[Lepikhin et~al.(2020)Lepikhin, Lee, Xu, Chen, Firat, Huang, Krikun,
  Shazeer, and Chen]{lepikhin2020gshard}
Dmitry Lepikhin, HyoukJoong Lee, Yuanzhong Xu, Dehao Chen, Orhan Firat, Yanping
  Huang, Maxim Krikun, Noam Shazeer, and Zhifeng Chen.
\newblock Gshard: Scaling giant models with conditional computation and
  automatic sharding.
\newblock \emph{arXiv preprint arXiv:2006.16668}, 2020.

\bibitem[Levesque(2011)]{wnli}
Hector~J. Levesque.
\newblock The winograd schema challenge.
\newblock In \emph{AAAI Spring Symposium: Logical Formalizations of Commonsense
  Reasoning}. AAAI, 2011.
\newblock URL
  \url{http://dblp.uni-trier.de/db/conf/aaaiss/aaaiss2011-6.html#Levesque11}.

\bibitem[Liu et~al.(2019{\natexlab{a}})Liu, Gardner, Belinkov, Peters, and
  Smith]{trans_context}
Nelson~F. Liu, Matt Gardner, Yonatan Belinkov, Matthew~E. Peters, and Noah~A.
  Smith.
\newblock Linguistic knowledge and transferability of contextual
  representations.
\newblock \emph{CoRR}, abs/1903.08855, 2019{\natexlab{a}}.
\newblock URL \url{http://arxiv.org/abs/1903.08855}.

\bibitem[Liu et~al.(2019{\natexlab{b}})Liu, He, Chen, and
  Gao]{mtl_bert_liu2019}
Xiaodong Liu, Pengcheng He, Weizhu Chen, and Jianfeng Gao.
\newblock Multi-task deep neural networks for natural language understanding.
\newblock \emph{CoRR}, abs/1901.11504, 2019{\natexlab{b}}.
\newblock URL \url{http://arxiv.org/abs/1901.11504}.

\bibitem[Liu et~al.(2020)Liu, Cheng, He, Chen, Wang, Poon, and
  Gao]{liu2020adversarial}
Xiaodong Liu, Hao Cheng, Pengcheng He, Weizhu Chen, Yu~Wang, Hoifung Poon, and
  Jianfeng Gao.
\newblock Adversarial training for large neural language models, 2020.

\bibitem[Liu et~al.(2019{\natexlab{c}})Liu, Ott, Goyal, Du, Joshi, Chen, Levy,
  Lewis, Zettlemoyer, and Stoyanov]{roberta}
Yinhan Liu, Myle Ott, Naman Goyal, Jingfei Du, Mandar Joshi, Danqi Chen, Omer
  Levy, Mike Lewis, Luke Zettlemoyer, and Veselin Stoyanov.
\newblock Roberta: {A} robustly optimized {BERT} pretraining approach.
\newblock \emph{CoRR}, abs/1907.11692, 2019{\natexlab{c}}.
\newblock URL \url{http://arxiv.org/abs/1907.11692}.

\bibitem[McCann et~al.(2018)McCann, Keskar, Xiong, and
  Socher]{mccann2018natural}
Bryan McCann, Nitish~Shirish Keskar, Caiming Xiong, and Richard Socher.
\newblock The natural language decathlon: Multitask learning as question
  answering.
\newblock \emph{arXiv preprint arXiv:1806.08730}, 2018.

\bibitem[Merchant et~al.(2020)Merchant, Rahimtoroghi, Pavlick, and
  Tenney]{merchant2020happens}
Amil Merchant, Elahe Rahimtoroghi, Ellie Pavlick, and Ian Tenney.
\newblock What happens to bert embeddings during fine-tuning?
\newblock \emph{arXiv preprint arXiv:2004.14448}, 2020.

\bibitem[Nguyen et~al.(2020)Nguyen, Vu, and Nguyen]{nguyen2020bertweet}
Dat~Quoc Nguyen, Thanh Vu, and Anh~Tuan Nguyen.
\newblock Bertweet: A pre-trained language model for english tweets.
\newblock \emph{arXiv preprint arXiv:2005.10200}, 2020.

\bibitem[Peng et~al.(2020)Peng, Schwartz, Li, and
  Smith]{peng-etal-2020-mixture}
Hao Peng, Roy Schwartz, Dianqi Li, and Noah~A. Smith.
\newblock A mixture of h - 1 heads is better than h heads.
\newblock In \emph{Proceedings of the 58th Annual Meeting of the Association
  for Computational Linguistics}, pp.\  6566--6577, Online, July 2020.
  Association for Computational Linguistics.
\newblock \doi{10.18653/v1/2020.acl-main.587}.
\newblock URL \url{https://www.aclweb.org/anthology/2020.acl-main.587}.

\bibitem[Perez et~al.(2018)Perez, Strub, de~Vries, Dumoulin, and
  Courville]{Perez2018FiLMVR}
Ethan Perez, Florian Strub, Harm de~Vries, Vincent Dumoulin, and Aaron~C.
  Courville.
\newblock Film: Visual reasoning with a general conditioning layer.
\newblock In \emph{AAAI}, 2018.

\bibitem[Peters et~al.(2018)Peters, Neumann, Zettlemoyer, and
  Yih]{word_context}
Matthew~E. Peters, Mark Neumann, Luke Zettlemoyer, and Wen{-}tau Yih.
\newblock Dissecting contextual word embeddings: Architecture and
  representation.
\newblock \emph{CoRR}, abs/1808.08949, 2018.
\newblock URL \url{http://arxiv.org/abs/1808.08949}.

\bibitem[Phang et~al.(2018)Phang, F{\'{e}}vry, and
  Bowman]{DBLP:journals/corr/abs-1811-01088}
Jason Phang, Thibault F{\'{e}}vry, and Samuel~R. Bowman.
\newblock Sentence encoders on {STILTs}: Supplementary training on intermediate
  labeled-data tasks.
\newblock \emph{CoRR}, abs/1811.01088, 2018.
\newblock URL \url{http://arxiv.org/abs/1811.01088}.

\bibitem[Poliak et~al.(2018)Poliak, Haldar, Rudinger, Hu, Pavlick, White, and
  Van~Durme]{wic}
Adam Poliak, Aparajita Haldar, Rachel Rudinger, J.~Edward Hu, Ellie Pavlick,
  Aaron~Steven White, and Benjamin Van~Durme.
\newblock Collecting diverse natural language inference problems for sentence
  representation evaluation.
\newblock In \emph{Proceedings of the 2018 Conference on Empirical Methods in
  Natural Language Processing}, pp.\  67--81, Brussels, Belgium,
  October-November 2018. Association for Computational Linguistics.
\newblock \doi{10.18653/v1/D18-1007}.
\newblock URL \url{https://www.aclweb.org/anthology/D18-1007}.

\bibitem[Ponti et~al.(2021)Ponti, Vuli{\'c}, Cotterell, Parovic, Reichart, and
  Korhonen]{ponti-etal-2021-parameter}
Edoardo~M. Ponti, Ivan Vuli{\'c}, Ryan Cotterell, Marinela Parovic, Roi
  Reichart, and Anna Korhonen.
\newblock Parameter space factorization for zero-shot learning across tasks and
  languages.
\newblock \emph{Transactions of the Association for Computational Linguistics},
  9:\penalty0 410--428, 2021.
\newblock \doi{10.1162/tacl_a_00374}.
\newblock URL \url{https://aclanthology.org/2021.tacl-1.25}.

\bibitem[Pruksachatkun et~al.(2020)Pruksachatkun, Phang, Liu, Htut, Zhang,
  Pang, Vania, Kann, and Bowman]{pruksachatkun2020intermediate}
Yada Pruksachatkun, Jason Phang, Haokun Liu, Phu~Mon Htut, Xiaoyi Zhang,
  Richard~Yuanzhe Pang, Clara Vania, Katharina Kann, and Samuel~R Bowman.
\newblock Intermediate-task transfer learning with pretrained models for
  natural language understanding: When and why does it work?
\newblock \emph{arXiv preprint arXiv:2005.00628}, 2020.

\bibitem[Radford et~al.(2018)Radford, Narasimhan, Salimans, and
  Sutskever]{Radford2018ImprovingLU}
Alec Radford, Karthik Narasimhan, Tim Salimans, and Ilya Sutskever.
\newblock Improving language understanding by generative pre-training.
\newblock 2018.

\bibitem[Raffel et~al.(2019)Raffel, Shazeer, Roberts, Lee, Narang, Matena,
  Zhou, Li, and Liu]{t5}
Colin Raffel, Noam Shazeer, Adam Roberts, Katherine Lee, Sharan Narang, Michael
  Matena, Yanqi Zhou, Wei Li, and Peter~J. Liu.
\newblock Exploring the limits of transfer learning with a unified text-to-text
  transformer, 2019.

\bibitem[Rajpurkar et~al.(2016{\natexlab{a}})Rajpurkar, Zhang, Lopyrev, and
  Liang]{squad}
Pranav Rajpurkar, Jian Zhang, Konstantin Lopyrev, and Percy Liang.
\newblock {SQ}u{AD}: 100,000+ questions for machine comprehension of text.
\newblock In \emph{Proceedings of the 2016 Conference on Empirical Methods in
  Natural Language Processing}, pp.\  2383--2392, Austin, Texas, November
  2016{\natexlab{a}}. Association for Computational Linguistics.
\newblock \doi{10.18653/v1/D16-1264}.
\newblock URL \url{https://www.aclweb.org/anthology/D16-1264}.

\bibitem[Rajpurkar et~al.(2016{\natexlab{b}})Rajpurkar, Zhang, Lopyrev, and
  Liang]{squad_rajpurkar2016}
Pranav Rajpurkar, Jian Zhang, Konstantin Lopyrev, and Percy Liang.
\newblock {SQ}u{AD}: 100,000+ questions for machine comprehension of text.
\newblock In \emph{Proceedings of the 2016 Conference on Empirical Methods in
  Natural Language Processing}, pp.\  2383--2392, Austin, Texas, November
  2016{\natexlab{b}}. Association for Computational Linguistics.
\newblock \doi{10.18653/v1/D16-1264}.

\bibitem[Reichart et~al.(2008)Reichart, Tomanek, Hahn, and
  Rappoport]{reichart2008multi}
Roi Reichart, Katrin Tomanek, Udo Hahn, and Ari Rappoport.
\newblock Multi-task active learning for linguistic annotations.
\newblock In \emph{Proceedings of ACL-08: HLT}, pp.\  861--869, 2008.

\bibitem[Ruder(2017)]{Ruder2017AnOO}
Sebastian Ruder.
\newblock An overview of multi-task learning in deep neural networks.
\newblock \emph{ArXiv}, abs/1706.05098, 2017.

\bibitem[Sener \& Koltun(2018)Sener and
  Koltun]{DBLP:journals/corr/abs-1810-04650}
Ozan Sener and Vladlen Koltun.
\newblock Multi-task learning as multi-objective optimization.
\newblock \emph{CoRR}, abs/1810.04650, 2018.
\newblock URL \url{http://arxiv.org/abs/1810.04650}.

\bibitem[Serrà et~al.(2018)Serrà, Suris, Miron, and
  Karatzoglou]{DBLP:conf/icml/SerraSMK18}
Joan Serrà, Didac Suris, Marius Miron, and Alexandros Karatzoglou.
\newblock Overcoming catastrophic forgetting with hard attention to the task.
\newblock In \emph{ICML}, pp.\  4555--4564, 2018.
\newblock URL \url{http://proceedings.mlr.press/v80/serra18a.html}.

\bibitem[Socher et~al.(2013)Socher, Perelygin, Wu, Chuang, Manning, Ng, and
  Potts]{sst}
Richard Socher, Alex Perelygin, Jean Wu, Jason Chuang, Christopher~D. Manning,
  Andrew Ng, and Christopher Potts.
\newblock Recursive deep models for semantic compositionality over a sentiment
  treebank.
\newblock In \emph{Proceedings of the 2013 Conference on Empirical Methods in
  Natural Language Processing}, pp.\  1631--1642, Seattle, Washington, USA,
  October 2013. Association for Computational Linguistics.
\newblock URL \url{https://www.aclweb.org/anthology/D13-1170}.

\bibitem[Standley et~al.(2019)Standley, Zamir, Chen, Guibas, Malik, and
  Savarese]{DBLP:journals/corr/abs-1905-07553}
Trevor Standley, Amir~Roshan Zamir, Dawn Chen, Leonidas~J. Guibas, Jitendra
  Malik, and Silvio Savarese.
\newblock Which tasks should be learned together in multi-task learning?
\newblock \emph{CoRR}, abs/1905.07553, 2019.
\newblock URL \url{http://arxiv.org/abs/1905.07553}.

\bibitem[Stickland et~al.(2019)Stickland, Murray, someone, and
  someone]{pmlr-v97-stickland19a}
Asa~Cooper Stickland, Iain Murray, someone, and someone.
\newblock {BERT} and {PAL}s: Projected attention layers for efficient
  adaptation in multi-task learning.
\newblock volume~97 of \emph{Proceedings of Machine Learning Research}, pp.\
  5986--5995, Long Beach, California, USA, 09--15 Jun 2019. PMLR.
\newblock URL \url{http://proceedings.mlr.press/v97/stickland19a.html}.

\bibitem[Tay et~al.(2020)Tay, Zhao, Bahri, Metzler, and Juan]{tay2020hypergrid}
Yi~Tay, Zhe Zhao, Dara Bahri, Donald Metzler, and Da-Cheng Juan.
\newblock Hypergrid: Efficient multi-task transformers with grid-wise
  decomposable hyper projections.
\newblock \emph{arXiv preprint arXiv:2007.05891}, 2020.

\bibitem[Tenney et~al.(2019{\natexlab{a}})Tenney, Das, and
  Pavlick]{bert_classicnlp}
Ian Tenney, Dipanjan Das, and Ellie Pavlick.
\newblock {BERT} rediscovers the classical {NLP} pipeline.
\newblock \emph{CoRR}, abs/1905.05950, 2019{\natexlab{a}}.
\newblock URL \url{http://arxiv.org/abs/1905.05950}.

\bibitem[Tenney et~al.(2019{\natexlab{b}})Tenney, Xia, Chen, Wang, Poliak,
  McCoy, Kim, Durme, Bowman, Das, and Pavlick]{context_prob}
Ian Tenney, Patrick Xia, Berlin Chen, Alex Wang, Adam Poliak, R.~Thomas McCoy,
  Najoung Kim, Benjamin~Van Durme, Samuel~R. Bowman, Dipanjan Das, and Ellie
  Pavlick.
\newblock What do you learn from context? probing for sentence structure in
  contextualized word representations.
\newblock \emph{CoRR}, abs/1905.06316, 2019{\natexlab{b}}.
\newblock URL \url{http://arxiv.org/abs/1905.06316}.

\bibitem[Trischler et~al.(2017)Trischler, Wang, Yuan, Harris, Sordoni, Bachman,
  and Suleman]{newsqa}
Adam Trischler, Tong Wang, Xingdi Yuan, Justin Harris, Alessandro Sordoni,
  Philip Bachman, and Kaheer Suleman.
\newblock {N}ews{QA}: A machine comprehension dataset.
\newblock In \emph{Proceedings of the 2nd Workshop on Representation Learning
  for {NLP}}, pp.\  191--200, Vancouver, Canada, August 2017. Association for
  Computational Linguistics.
\newblock \doi{10.18653/v1/W17-2623}.
\newblock URL \url{https://www.aclweb.org/anthology/W17-2623}.

\bibitem[Vaswani et~al.(2017)Vaswani, Shazeer, Parmar, Uszkoreit, Jones, Gomez,
  Kaiser, and Polosukhin]{transformer}
Ashish Vaswani, Noam Shazeer, Niki Parmar, Jakob Uszkoreit, Llion Jones,
  Aidan~N. Gomez, Lukasz Kaiser, and Illia Polosukhin.
\newblock Attention is all you need.
\newblock \emph{CoRR}, abs/1706.03762, 2017.
\newblock URL \url{http://arxiv.org/abs/1706.03762}.

\bibitem[von Oswald et~al.(2020)von Oswald, Henning, Grewe, and
  Sacramento]{Oswald2020Continual}
Johannes von Oswald, Christian Henning, Benjamin~F. Grewe, and João
  Sacramento.
\newblock Continual learning with hypernetworks.
\newblock In \emph{International Conference on Learning Representations}, 2020.
\newblock URL \url{https://openreview.net/forum?id=SJgwNerKvB}.

\bibitem[Wang et~al.(2018)Wang, Singh, Michael, Hill, Levy, and
  Bowman]{wang-etal-2018-glue}
Alex Wang, Amanpreet Singh, Julian Michael, Felix Hill, Omer Levy, and Samuel
  Bowman.
\newblock {GLUE}: A multi-task benchmark and analysis platform for natural
  language understanding.
\newblock In \emph{Proceedings of the 2018 {EMNLP} Workshop {B}lackbox{NLP}:
  Analyzing and Interpreting Neural Networks for {NLP}}, pp.\  353--355,
  Brussels, Belgium, November 2018. Association for Computational Linguistics.
\newblock \doi{10.18653/v1/W18-5446}.
\newblock URL \url{https://www.aclweb.org/anthology/W18-5446}.

\bibitem[Wang et~al.(2019{\natexlab{a}})Wang, Hula, Xia, Pappagari, McCoy,
  Patel, Kim, Tenney, Huang, Yu, Jin, Chen, {Van Durme}, Grave, Pavlick, and
  Bowman]{wang-acl-19}
Alex Wang, Jan Hula, Patrick Xia, Raghavendra Pappagari, R.~Thomas McCoy, Roma
  Patel, Najoung Kim, Ian Tenney, Yinghui Huang, Katherin Yu, Shuning Jin,
  Berlin Chen, Benjamin {Van Durme}, Edouard Grave, Ellie Pavlick, and
  Samuel~R. Bowman.
\newblock Can you tell me how to get past sesame street? sentence-level
  pretraining beyond language modeling.
\newblock In \emph{Proceedings of the Annual Meeting of the Association for
  Computational Linguistics (ACL)}, 2019{\natexlab{a}}.

\bibitem[Wang et~al.(2019{\natexlab{b}})Wang, Pruksachatkun, Nangia, Singh,
  Michael, Hill, Levy, and Bowman]{superglue}
Alex Wang, Yada Pruksachatkun, Nikita Nangia, Amanpreet Singh, Julian Michael,
  Felix Hill, Omer Levy, and Samuel~R. Bowman.
\newblock Superglue: {A} stickier benchmark for general-purpose language
  understanding systems.
\newblock \emph{CoRR}, abs/1905.00537, 2019{\natexlab{b}}.
\newblock URL \url{http://arxiv.org/abs/1905.00537}.

\bibitem[Warstadt et~al.(2018)Warstadt, Singh, and Bowman]{cola}
Alex Warstadt, Amanpreet Singh, and Samuel~R. Bowman.
\newblock Neural network acceptability judgments.
\newblock \emph{CoRR}, abs/1805.12471, 2018.
\newblock URL \url{http://arxiv.org/abs/1805.12471}.

\bibitem[Williams et~al.(2018)Williams, Nangia, and Bowman]{mnli}
Adina Williams, Nikita Nangia, and Samuel Bowman.
\newblock A broad-coverage challenge corpus for sentence understanding through
  inference.
\newblock In \emph{Proceedings of the 2018 Conference of the North {A}merican
  Chapter of the Association for Computational Linguistics: Human Language
  Technologies, Volume 1 (Long Papers)}, pp.\  1112--1122, New Orleans,
  Louisiana, June 2018. Association for Computational Linguistics.
\newblock \doi{10.18653/v1/N18-1101}.
\newblock URL \url{https://www.aclweb.org/anthology/N18-1101}.

\bibitem[Wolf et~al.(2019)Wolf, Debut, Sanh, Chaumond, Delangue, Moi, Cistac,
  Rault, Louf, Funtowicz, and Brew]{huggingface}
Thomas Wolf, Lysandre Debut, Victor Sanh, Julien Chaumond, Clement Delangue,
  Anthony Moi, Pierric Cistac, Tim Rault, R{\'{e}}mi Louf, Morgan Funtowicz,
  and Jamie Brew.
\newblock Huggingface's transformers: State-of-the-art natural language
  processing.
\newblock \emph{CoRR}, abs/1910.03771, 2019.
\newblock URL \url{http://arxiv.org/abs/1910.03771}.

\bibitem[Wu et~al.(2020)Wu, Zhang, and Ré]{Wu2020Understanding}
Sen Wu, Hongyang~R. Zhang, and Christopher Ré.
\newblock Understanding and improving information transfer in multi-task
  learning.
\newblock In \emph{International Conference on Learning Representations}, 2020.
\newblock URL \url{https://openreview.net/forum?id=SylzhkBtDB}.

\bibitem[Yang et~al.(2018)Yang, Qi, Zhang, Bengio, Cohen, Salakhutdinov, and
  Manning]{hotpotqa}
Zhilin Yang, Peng Qi, Saizheng Zhang, Yoshua Bengio, William Cohen, Ruslan
  Salakhutdinov, and Christopher~D. Manning.
\newblock {H}otpot{QA}: A dataset for diverse, explainable multi-hop question
  answering.
\newblock In \emph{Proceedings of the 2018 Conference on Empirical Methods in
  Natural Language Processing}, pp.\  2369--2380, Brussels, Belgium,
  October-November 2018. Association for Computational Linguistics.
\newblock \doi{10.18653/v1/D18-1259}.
\newblock URL \url{https://www.aclweb.org/anthology/D18-1259}.

\bibitem[Yu et~al.(2020)Yu, Kumar, Gupta, Levine, Hausman, and
  Finn]{yu2020gradient}
Tianhe Yu, Saurabh Kumar, Abhishek Gupta, Sergey Levine, Karol Hausman, and
  Chelsea Finn.
\newblock Gradient surgery for multi-task learning.
\newblock \emph{arXiv preprint arXiv:2001.06782}, 2020.

\bibitem[Zhang et~al.(2018)Zhang, Liu, Liu, Gao, Duh, and Durme]{record}
Sheng Zhang, Xiaodong Liu, Jingjing Liu, Jianfeng Gao, Kevin Duh, and
  Benjamin~Van Durme.
\newblock Record: Bridging the gap between human and machine commonsense
  reading comprehension.
\newblock \emph{CoRR}, abs/1810.12885, 2018.
\newblock URL \url{http://arxiv.org/abs/1810.12885}.

\bibitem[Zhang \& Yang(2017)Zhang and Yang]{DBLP:journals/corr/ZhangY17aa}
Yu~Zhang and Qiang Yang.
\newblock A survey on multi-task learning.
\newblock \emph{CoRR}, abs/1707.08114, 2017.
\newblock URL \url{http://arxiv.org/abs/1707.08114}.

\end{thebibliography}
\bibliographystyle{iclr2021_conference}

\clearpage

\appendix

\section{Appendix}

\subsection{Summary of Acronyms}
\label{append:acronyms}

Acronyms of datasets and descriptions can be found below in section \ref{append:datasets}.

\begin{table}[ht]
\caption{\small List of acronyms used in this paper.}
\begin{center}
\scriptsize
\begin{tabular}{|l|l|}
	\hline 
		\textbf{Acronym}   & \textbf{Description} \\
		\hline
		ARLM      & Autoregressive Language Models \\
		CA-MTL    & Conditional Adaptive Multi-Task Learning: our architecture \\
		CFF       & Conditional Feed-Forward: a feed-forward layer modulated by a conditioning vector \\
		CLN       & Conditional Layer Normalization in section \ref{sec:cond_layer_norm} \\
		EDM       & Evolutionary  Data  Measures \citep{collins-etal-2018-evolutionary}: a task difficulty estimate \\
		GLUE      & General Language Understanding Evaluation \cite{wang-etal-2018-glue}: a benchmark with multiple datasets \\
		QA        & Question Answering \\
		MT        & Multi-Task \\
		MTAL      & Multi-Task Active Learning: finding the most informative instance for multiple learners (or models) \\
		MLM       & Masked Language Model: BERT \cite{bert} is an example of an MLM \\
		MTL       & Multi-Task Learning: "learning tasks in parallel while using a shared representation" \citep{mtl_caruana1997} \\
		MRQA      & Machine Reading for Question Answering \cite{fisch2019mrqa}: a benchmark 
		with multiple datasets  \\
		NER       & Named Entity Recognition \\
		NLP       & Natural Language Processing \\
		SOTA      & State of the art \\
		ST        & Single Task fine-tuning: all weights are typically updated  \\
		ST-A      & ST with Adapter modules: one adapter per task is trained and pretrained weights are optionally updated \\
		
    \hline
\end{tabular}
\end{center}

\label{table:acronyms}
\end{table}

\subsection{Uncertainty Sampling: Algorithm and Additional Results}
\label{append:uncert_sampling}

\begin{figure}[hbt!]
    \begin{center}
        \scalebox{.6}{\input{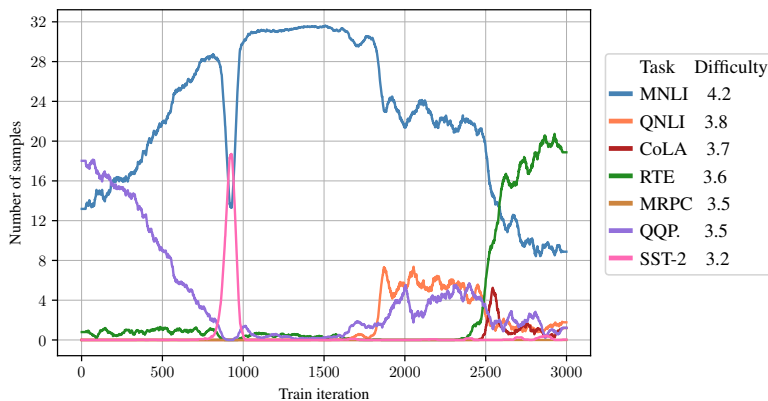}}
    \end{center}
\caption{\small Task composition of MT-Uncertainty sampling and estimated task difficulty using EDM: number of training samples per task at each iteration for batch size of 32. The occurrence of first peaks and estimated difficulty follow the same order: From highest to lowest: MNLI $>$ CoLA $>$ RTE $>$ QQP $=$ MRPC $>$ SST-2.}
\label{fig:task_diff}
\end{figure}

\begin{minipage}{0.9\linewidth}
\begin{algorithm}[H]
\caption{Multi-task Uncertainty Sampling}
    \SetAlgoLined
    \DontPrintSemicolon
    \label{alg:uncert_alg}  
    \SetCustomAlgoRuledWidth{1.2cm}  
    \KwInput{Training data  \( D_{t}\) for task \( t \in [1,\dotsc,T] \); batch size \(b\); \(C_t\) possible output classes for task $t$; \(f:=f_{\phi(\textbf{z}_i),\theta_i}\) our model with weights $\phi,\theta_i$;}
    \KwOutput{\(\mathcal{B}'\) - multi-task batch of size \(b\)}
    \(\mathcal{B} \gets \emptyset\)\;
    \For{$t\gets1$ \KwTo \(T\)}{
        Generate \( \textbf{x}_t := \{x_{t, 1}, \dotsc ,x_{t, b}\} \mytilde D_t\)\\
        \For{$i\gets1$ \KwTo \(b\)}{
            \(\mathcal{H}_{t,i} \gets -\sum_{c=1}^{C_i} p_c(f(x_{t,i}))\,\log\,p_c(f(x_{t,i}))\) \Comment*[r]{Entropy of each sample} \;
        }
        Compute \(\bar{\mathcal{H}}_t \gets \frac{1}{b} \sum_{\text{x} \in \textbf{x}_i} \mathcal{H}_{t,i}\) \Comment*[r]{Average entropy for task $t$} \;
        Compute \(H'_t \gets -\sum_{c=1}^{C_t} \frac{1}{C_t}\,\log \bigg[\frac{1}{C_t}\bigg]\) \Comment*[r]{Max entropy (uniform distribution)}\;
        \(\mathcal{B} \gets \mathcal{B} \cup \textbf{x}_t \)\ and \(D_t \gets D_t \setminus \textbf{x}_t \)\;
        \If{ \(D_t = \emptyset\)}{
            Reload \(D_t\)
        }
        \For{$i\gets1$ \KwTo \(b\)}{
            Compute: \(\mathcal{U}_{t,i} \gets \mathcal{H}_{t,i}/H'_t\) \Comment*[r]{Uncertainty normalized with max entropy}
        }
    }
    Compute $\hat{\mathcal{H}} \gets \max_{i \in \{1,\dotsc,T\}} [\bar{\mathcal{H}}_t]$ \Comment*[r]{Entropy of task with highest average entropy}
    Update \(\mathcal{U}_{t,i} \gets \mathcal{U}_{t,i}/\hat{\mathcal{H}}\) \Comment*[r]{Normalize each sample's uncertainty measure}
    \(\mathcal{B}' \gets \topb(\{\mathcal{U}_{t,i}| t \in [1,\dotsc,T], i \in [1,\dotsc,b]\})\) \Comment*[r]{$b$ samples w/ highest uncertainty}
    \KwReturn{With \(\mathcal{B}'\), solve eq. \ref{eq:mt_obj} with gradient descent; updated model $f$}
\end{algorithm}
\end{minipage}
\clearpage
An advantage of our MT-Uncertainty Sampling approach is its ability to manage task difficulty. This is highlighted in Figure \ref{fig:task_diff}. In this experiment, we estimated task difficulty using the Evolutionary Data Measures (EDM)\footnote{https://github.com/Wluper/edm} proposed by \citet{collins-etal-2018-evolutionary}. The task difficulty estimate relies on multiple dataset statistics such as the data size, class diversity, class balance and class interference. Interestingly, estimated task difficulty correlates with the first instance that the selection of a specific task occurs. Supposing that QNLI is an outlier, we notice that peaks in the data occur whenever tasks are first selected by MT Uncertainty sampling. This process follows the following order: 1. MNLI 2. CoLA 3. RTE 4. QQP 5. MRPC 6.SST-2, which is the order from highest task difficulty to lowest task difficulty using EDM. As opposed to Curriculum Learning \citep{bengio2009curriculum}, MT-Uncertainty dynamically prioritizes the most difficult tasks. As also discovered in MTL vision work \citep{Guo_2018_ECCV}, this type of prioritization on more difficult tasks may explain MT-Uncertainty's improved performance over other task selection methods. In MTL, heuristics to balance tasks during training is typically done by weighting each task's loss differently. We see here how MT-Uncertainty is able to prioritize task difficulty.

While the EDM difficulty measure is shown to correlate well with model performance, it lacks precision. As reported in \citet{collins-etal-2018-evolutionary}, the average score achieved on the Yahoo Answers dataset is 69.9\% and its difficulty is 4.51. The average score achieved on Yelp Full is 56.8\%, 13.1\% less than Yahoo Answers and its difficulty is 4.42. The authors mention that ``This indicates that the difficulty measure in its current incarnation may be more effective at assigning a class of difficulty to datasets, rather than a regression-like value''.

\subsection{Other Related Work}
\textbf{Multi-Tasking in NLP and other fields.}
MTL weight sharing algorithms such as Mixture-of-Experts (MoE) have found success in NLP \citep{lepikhin2020gshard}. CA-MTL can complement MoE since the Transformers multi-headed attention can be seen as a form of MoE \citep{peng-etal-2020-mixture}. In Vision, MTL can also improve with optimization \citep{DBLP:journals/corr/abs-1810-04650} or gradient-based approaches \citep{DBLP:journals/corr/abs-1711-02257,yu2020gradient}.

\textbf{Active Learning, Task Selection and Sampling.}
\citet{ikhwantri2018multi} examined multi-task active learning for neural semantic role labeling in a low resource setting, using entity recognition as the sole auxiliary task. They used uncertainty sampling for active learning and found that 12\% less data could be used compared to passive learning. 
\citet{reichart2008multi} has examined different active learning techniques for the two task annotation scenario, focusing on named entity recognition and syntactic parse tree annotations. 
In contrast, here we examine the larger scale data regime, the modularization of a multi-task neural architecture, and the many task ($\gg$ 2) setting among other differences. Other than MTAL \citep{reichart2008multi,ikhwantri2018multi}, \cite{DBLP:journals/corr/KendallGC17} leveraged model uncertainty to balance MTL losses but not to select tasks as is proposed here.


\subsection{Zero-Shot Results on SciTail and SNLI}
\label{append:newtask_embed_choice}
Before testing models on domain adaptation in section \ref{sec:new_tasks},  we ran zero-shot evaluations on the development set of SciTail and SNLI. Table \ref{table:data_zero_shot} outlines $\text{8-task CA-MTL}_{\text{BERT-BASE}}$'s zero-shot transfer abilities when  pretrained on GLUE with our MTL approach. We expand the task embedding layer to accommodate an extra task and explore various embedding initialization. We found that reusing STS-B and MRPC task embeddings worked best for SciTail and SNLI respectively.

\begin{table}[hbt]
\caption{\small CA-MTL is flexible and extensible to new tasks. However, CA-MTL is sensitive to the new task's embedding. We tested multiple task embeddings that worked best on either SciTail or SNLI by checking performance in a zero shot setting or using 0\% of the data.}
\begin{center}
\small
\begin{tabular}{|l|c|c|c|c|}
	\hline 
		Initialization of new & SciTail     & SNLI       \\
		task embedding layer  & 0\% of data & 0\% of data \\
		\hline
        CoLA's embeddings  & 43.0 & 34.0 \\
        MNLI's embeddings  & 24.2 & 33.0 \\
        MRPC's embeddings  & 34.5 & \textbf{45.5} \\
        STS-B's embeddings  & \textbf{46.9} & 33.2 \\
        SST-2's embeddings  & 25.8 & 34.2 \\
        QQP's embeddings   & 31.7 & 37.3  \\
        QNLI's embeddings   & 32.0 & 38.0 \\
        RTE's embeddings   & 32.3 & 40.6  \\
        WNLI's embeddings   & 29.0 & 30.4 \\
        Average  & 28.7 &  37.7 \\
        Random initialization & 46.8 & 34.0 \\
        Xavier initialization & 29.8 &  37.6 \\
    \hline
\end{tabular}
\end{center}

\label{table:data_zero_shot}
\end{table}

\subsection{More Experimental Details}
\label{append:more_xp_details}
We used a batch size of 32 and a seed of 12 in all experiments. We used Adam \citep{Kingma2015AdamAM} as the optimizer with a learning rate of 2e-5. We applied a learning rate decay with warm up over the first 10\% of the training steps. Unless otherwise specified, we used 5 epochs, a seed of 12 and a sequence length of 128. Additional details are outlined in section . Our data prepossessing and linear decoder heads are the same as in \citet{bert}. We used the same dropout rate of 0.1 in all layers. To run our experiments, we used either four NVIDIA P100 GPU for base models or four NVIDIA V100 GPU for larger ones. We did not perform parameter search. We do not use ensemble of models or task-specific tricks \citep{bert,mtl_bert_liu2019,mtl_bert_clark2019}. All models are either 12 Transformer layers for BASE and 24 Transformer layers for LARGE. Apart from CA-MTL, models trained in multi-task learning (BERT or RoBERTa without adapters) used random task sampling. For Table \ref{table:architectural_ablation} and Figure \ref{fig:data_ablation}, all BERT-based model have half their layers frozen (untrained) for a fair comparison of ablation results. For the 24-task MTL and CA-MTL models in Tables \ref{table:24-task-camtl} and \ref{table:glue-sota}, we increased the input sequence length to 256 and used 8 epochs. 

\subsection{The Direct Sum Operator}
\label{append:directsum}
In section \ref{sec:cond_attn_mat}, we used the direct sum operator $\oplus$. This operation allows us to create a block diagonal matrix. The direct sum of a matrix $A \in \mathbb{R}^{n\times m}$ and $B \in \mathbb{R}^{p\times q}$ results in a matrix of size  $(m + p) \times (n + q)$, defined as:\newline
\begin{center}
$\mathbf{A} \oplus \mathbf{B} =
  \begin{bmatrix} \mathbf{A} & \boldsymbol{0} \\ \boldsymbol{0} & \mathbf{B} \end{bmatrix} =
  \begin{bmatrix}
     a_{11} & \cdots & a_{1n} &      0 & \cdots &      0 \\
     \vdots & \ddots & \vdots & \vdots & \ddots & \vdots \\
    a_{m 1} & \cdots & a_{mn} &      0 & \cdots &      0 \\
          0 & \cdots &      0 & b_{11} & \cdots &  b_{1q} \\
     \vdots & \ddots & \vdots & \vdots & \ddots & \vdots \\
          0 & \cdots &      0 & b_{p1} & \cdots &  b_{pq}
  \end{bmatrix}$
\end{center}

\subsection{Baselines and Other Experimental Results}
\label{append:more_xp_results}
In this section, we present our baseline results for BERT, RoBERTa, CA-MTL as well as other models. Our single task results (ST) that we ran ourselves surpass other paper's reported scores in Table \ref{table:glue_baselines}. \cite{roberta} reports random seed median scores for RoBERTa. However, our RoBERTa ST baseline \textbf{matches or surpasses} the original paper's scores \textbf{4 out 7 times} on the development set when scores are comparable (QQP F1 and STS-B spearman are not reported).
\begin{table*}[h!]
\caption{\small F1 scores are reported for QQP/MRPC, Spearman's correlation for STS-B, accuracy on the matched/mismatch sets for MNLI, Matthew's correlation for CoLA and accuracy for other tasks. ST=Single Task, MTL=Multitask. *QNLI v1 (we report v2) **F1 score or Spearman's correlation is not reported. ***Unknown random seeds.
Results from:
$^{1}$\citet{pmlr-v97-stickland19a}
$^{2}$\cite{mtl_bert_liu2019}
$^{3}$\cite{DBLP:journals/corr/abs-1811-01088}
$^{4}$\cite{roberta}.
}
\label{table:glue_baselines}
\begin{center}
\scriptsize
\setlength{\tabcolsep}{2pt}
\begin{tabular}{|c|c|c|ccccccccc|c}
	\hline 
		\multirow{2}*{Method} & Total & Trained & \multicolumn{9}{c|}{GLUE}  \\
        & params & params/task & CoLA & MNLI & MRPC & QNLI & QQP & RTE & SST-2 & STS-B & Avg \\ \hline
        
        \hline
        \multicolumn{12}{|c|}{\textbf{Base Models --- Dev set Results}} \\
        \hline
        
        PALs+Anneal Samp.$^{1}$ & 1.13$\times$ & 12.5\% & -- & -- & -- & -- & -- & -- & -- & -- & 81.70 \\
        
        $\text{8-task CA-MTL}_{\text{BERT-BASE}}$ (ours) & 1.12$\times$ & 5.6\% & 60.9 & 82.7/83.1 & 88.9 & 90.7 & 90.3 & 79.1 & 91.9 & 88.8 & 84.03  \\
        
        \hline
        \multicolumn{12}{|c|}{\textbf{BERT LARGE Models --- Dev set Results}} \\
        \hline
        ST BERT-LARGE$^{2}$  & 9$\times$ & 100\% & 60.5 & 86.7/85.9 & 89.3 & 92.7* & 89.3 & 70.1 & 94.9 & 86.5 & 84.0 \\
        ST BERT-LARGE$^{3}$  & 9$\times$ & 100\% & 62.1 & 86.2/86.2 & 92.3 & 89.4  & 88.5 & 70.0 & 92.5 & 90.1 & 84.1 \\
        ST BERT-LARGE (ours) & 9$\times$ & 100\% & 63.6 & 86.5/86.0 & 91.4 & 91.0  & 88.5 & 70.2 & 94.7 & 88.2 & 84.5 \\
        $\text{24-task CA-MTL}_{\text{BERT-LARGE}}$ (ours) & 1.12$\times$    & 5.6\% & 63.8 & 86.3/86.0 & 92.9 & 93.4 & 88.1 & 84.5 & 94.5 & 90.3 & 86.6  \\
        \hline
        \multicolumn{12}{|c|}{\textbf{RoBERTa LARGE Models --- Dev set Results}} \\
        \hline
        RoBERTa-LARGE$^{4}$  & \multirow{2}*{9$\times$} & \multirow{2}*{100\%} & \multirow{2}*{68.0} & \multirow{2}*{90.2} & \multirow{2}*{90.9} & \multirow{2}*{94.7}  & \multirow{2}*{**} & \multirow{2}*{86.6} & \multirow{2}*{96.4} & \multirow{2}*{**} & \multirow{2}*{--} \\
        (Median 5 runs)***  & & & & & & & & & & &  \\
        ST RoBERTa-LARGE (ours)  & 9$\times$ & 100\% & 68.3 & 89.2/88.9 & 92.6 & 94.8 & 84.6 & 87.0 & 96.4 & 91.7 & 88.2 \\
        $\text{24-task CA-MTL}_{\text{RoBERTa-LARGE}}$ (ours) & 1.12$\times$ & 5.6\% & 69.7 & 89.4/89.3 & 93.9 & 94.9 & 88.8 & 91.0 & 96.2 & 91.0 & 89.4  \\
        
    \hline
\end{tabular}
\end{center}
\vspace{-.25cm}
\end{table*}

\subsection{Some Results on layer Freezing and with Full Block Attention.}
\label{append:freeze_attn_block}
All experiments in this section were run for only 5 epochs, exclusively on the GLUE dataset for the large BERT-based 8-task CA-MTL model. Results in Table \ref{table:glue_freeze} reveal that as we freeze more layers, performance tends to decrease. However, since we wanted to preserve as much pretrained knowledge as possible, we chose to keep at least 50\% of layers frozen. While this has slightly lowered our performance on 9 GLUE tasks, we believe that keeping as much of the original pretrained weights is beneficial when increasing the total number of tasks in MTL to 24 or more tasks. However, we did not explore this hypothesis more.

\begin{table*}[h!]
\caption{\small $\textbf{8-task CA-MTL}_\textbf{BERT-LARGE}$ (see section \ref{sec:adapters}) for various layer freezing configurations. F1 scores are reported for QQP/MRPC, Spearman's correlation for STS-B, accuracy on the matched/mismatch sets for MNLI, Matthew's correlation for CoLA and accuracy for other tasks. FBA = Full Block Attention
}
\label{table:glue_freeze}
\begin{center}
\scriptsize
\setlength{\tabcolsep}{2pt}
\begin{tabular}{|c|c|c|ccccccccc|c}
	\hline 
		\multirow{2}*{Method} & \% frozen & \# tasks & \multicolumn{9}{c|}{GLUE}  \\
                              & layers    & g.e ST & CoLA & MNLI & MRPC & QNLI & QQP & RTE & SST-2 & STS-B & Avg \\ \hline
        
        \hline
        \multicolumn{12}{|c|}{\textbf{LARGE Models --- Dev set Results}} \\
        \hline
        ST BERT-LARGE (ours) & 0\% & ---  & 63.6 & 86.5/86.0 & 91.4 & 91.0 & 88.5 & 70.2 & 93.1 & 88.2 & 84.3  \\
        $\text{CA-MTL}$      & 0\%  & 7 & 60.2 & 86.2/86.0 & 92.0 & 91.5 & 88.7 & 76.3 & 93.3 & 89.5 & 84.9  \\
        $\text{CA-MTL}$      & 25\% & 6 & 63.7 & 86.1/85.8 & 89.1 & 91.2 & 88.6 & 79.7 & 92.9 & 88.5 & 85.1  \\
        $\text{CA-MTL}$      & 50\% & 3 & 63.2 & 85.5/85.5 & 91.8 & 90.9 & 88.3 & 81.4 & 93.0 & 90.1 & 85.5  \\
        $\text{CA-MTL}$ FBA  & 50\% & 0 & 60.2 & 81.7/81.1 & 88.0 & 85.8 & 85.7 & 78.7 & 88.6 & 87.1 & 81.8  \\
        
    \hline
\end{tabular}
\end{center}
\vspace{-.25cm}
\end{table*}

\subsection{Dataset Description}
\label{append:datasets}

The datasets that were used for the domain adaptation experiments were SciTail\footnote{https://allenai.org/data/scitail; Leaderboard can be found at: https://leaderboard.allenai.org/scitail/submissions/public} and SNLI\footnote{https://nlp.stanford.edu/projects/snli/}. We \emph{jointly} trained a $\text{CA-MTL}_{\text{RoBERTa-LARGE}}$ model on 9 GLUE tasks, 8 Super-GLUE\footnote{https://super.gluebenchmark.com/tasks} tasks, 6 MRQA\footnote{https://github.com/mrqa/MRQA-Shared-Task-2019} tasks, and on WNUT2017\footnote{https://github.com/leondz/emerging\_entities\_17} \citep{derczynski-etal-2017-results}. 

\begin{table}[hbt]
\caption{\small GLUE \citep{wang-etal-2018-glue} dataset description. \\
References: $^1$\citet{cola}, $^2$\citet{sst}, $^3$\citet{mrpc},  $^4$\citet{sts-b}, $^5$\citet{mnli}, $^6$\citet{wang-etal-2018-glue}, $^7$\citet{wnli} }
\begin{center}
\small
\begin{tabular}{|l|l|l|l|l|}
	\hline 
		\textbf{Acronym} & \textbf{Corpus} & $\lvert\textbf{Train}\rvert$ & \textbf{Task} & \textbf{Domain}  \\
		\hline
		CoLA$^1$  & Corpus of Linguistic Acceptability  & 8.5K & acceptability & miscellaneous \\
		SST-2$^2$ & Stanford Sentiment Treebank & 67K & sentiment detection & movie reviews \\
		MRPC$^3$  & Microsoft Research Paraphrase Corpus & 3.7K & paraphrase detection & news \\
		STS-B$^4$ & Semantic Textual Similarity Benchmark & 7K & textual similarity & miscellaneous \\
		QQP   & Quora Question Pairs & 364K & paraphrase detection & online QA \\
		MNLI$^5$  & Multi-Genre NLI & 393K & inference &  miscellaneous \\
		RTE$^6$   & Recognition Textual Entailment & 2.5K & inference/entailment & news, Wikipedia \\
		WNLI$^7$  & Winograd NLI & 634 & coreference & fiction books \\

    \hline
\end{tabular}
\end{center}

\label{table:data}
\end{table}

All GLUE tasks are binary classification, except STS-B (regression) and MNLI (three classes). We used the same GLUE data preprocessing as in \citet{bert}.

\begin{table}[ht]
\caption{\small Super-GLUE \citep{superglue} dataset description. References: $^1$\citet{boolq}, $^2$\citet{cb}, $^3$\citet{copa},  $^4$\citet{multirc}, $^5$\citet{record}, $^6$\citet{superglue}, $^7$\citet{wic}, $^8$\citet{wnli} }
\begin{adjustbox}{width=\columnwidth,center}
\begin{tabular}{|l|l|l|l|l|}
	\hline 
		\textbf{Acronym} & \textbf{Corpus} & $\lvert\textbf{Train}\rvert$ & \textbf{Task} & \textbf{Domain}  \\
		\hline
		BoolQ$^1$   & Boolean Questions  & 9.4K & acceptability & Google queries, Wikipedia \\
		CB$^2$      & CommitmentBank & 250 & sentiment detection & miscellaneous \\
		COPA$^3$    & Choice of Plausible Alternatives & 400 & paraphrase detection & blogs, encyclopedia \\
		MultiRC$^4$ & Multi-Sentence Reading Comprehension & 5.1K & textual similarity & miscellaneous \\
		ReCoRD$^5$  & Reading Comprehension & 101K & paraphrase detection & news \\
		            & and Commonsense Reasoning & & & \\
		RTE$^6$     & Recognition Textual Entailment & 2.5K & inference & news, Wikipedia \\
		WiC$^7$     & Word-in-Context & 6K & word sense disambiguation & WordNet, VerbNet \\
		WSC$^8$     & Winograd Schema Challenge & 554 & coreference resolution & fiction books \\

    \hline
\end{tabular}
\end{adjustbox}
\end{table}

\begin{table}[H]
\caption{\small MRQA \citep{fisch2019mrqa} dataset description. References: $^1$\cite{squad}, $^2$\citet{newsqa}, $^3$\citet{triviaqa},  $^4$\citet{searchqa}, $^5$\citet{hotpotqa}, $^6$\citet{natq} }
\begin{center}
\footnotesize
\begin{tabular}{|l|l|l|l|l|}
	\hline 
		\textbf{Acronym} & \textbf{Corpus} & $\lvert\textbf{Train}\rvert$ & \textbf{Task} & \textbf{Domain}  \\
		\hline
		SQuAD$^1$             & Stanford QA Dataset & 86.6K  & crowdsourced questions & Wikipedia \\
		NewsQA$^2$            & NewsQA              & 74.2K  & crowdsourced questions & news \\
		TriviaQA$^3$          & TriviaQA            & 61.7K  & trivia QA & web snippets \\
		SearchQA$^4$          & SearchQA            & 117.4K & Jeopardy QA & web snippets \\
		HotpotQA$^5$          & HotpotQA            & 72.9K  & crowdsourced questions & Wikipedia \\
		Natural Questions$^6$ & Natural Questions   & 104.7K & search logs & Wikipedia \\
    \hline
\end{tabular}
\end{center}

\end{table}
SuperGLUE has a more diverse task format than GLUE, which is mostly limited to sentence and sentence-pair classification. We follow the same preprocessing procedure as in \citet{superglue}. All tasks are binary classification tasks, except CB (three classes). Also, WiC and WSC are span based classification tasks. We used the same modified MRQA dataset and preprocessing steps that were used in \citet{DBLP:journals/corr/abs-1907-10529}. All MRQA tasks are span prediction tasks  which seeks to identify start and end tokens of an answer span in the input text.

\begin{table}[H]
\caption{\small SNLI \citep{snli:emnlp2015} and SciTail \citep{Khot2018SciTaiLAT} datasets description.}
\begin{center}
\small
\setlength{\tabcolsep}{2pt}
\begin{tabular}{|l|l|l|l|l|}
	\hline 
		\textbf{Acronym} & \textbf{Corpus} & $\lvert\textbf{Train}\rvert$ & \textbf{Task} & \textbf{Domain}  \\
		\hline
		SNLI$^1$             & Stanford Natural Language Inference & 550.2k  & inference & human-written English sentence pairs \\
		SciTail$^2$            & Science and Entailment              & 23.5K  & entailment & Science question answering \\
    \hline
\end{tabular}
\end{center}
\end{table}

SNLI is a natural inference task where we predict three classes. Examples of three target labels are: Entailment, Contradiction, and Neutral (irrelevant). SciTail is a textual entailment dataset. The hypotheses in SciTail are created from multiple-choice science exams and the answer candidates (premise) are extracted from the web using information retrieval tools. SciTail is a binary true/false classification tasks that seeks to predict whether the premise entails the hypothesis. The two datasets are used only for domain adaptation in this study (see section \ref{append:newtask_embed_choice} for the details of our approach).
\clearpage

\end{document}